\documentclass{ifm-tr}
\usepackage{mathpazo}
\ifPDFTeX\else
  \defaultfontfeatures{Ligatures=TeX}% XCharter sets Extension=.otf for every later font
  \IfFontExistsTF{Palatino Linotype}{\setmainfont{Palatino Linotype}}{%
    \setmainfont{texgyrepagella}[Extension=.otf, UprightFont=*-regular, BoldFont=*-bold,
      ItalicFont=*-italic, BoldItalicFont=*-bolditalic]}
\fi

\usepackage{amsmath,amssymb,enumitem,url,float,wrapfig}
\usepackage{algorithm,algorithmic}
\usepackage{fontawesome5}
\newcommand{\titlesubtitlelist}{}

\renewcommand{\title}[1]{\def\titlelist{#1}}

\newcommand{\patchpartititle}[2]{%
  \patchcmd{\mymaketitle}{#1}{#2}{}{%
    \PackageError{parti-title}{The IFM title template changed}{Check title_layout.tex against ifm-tr.cls.}%
  }%
}
\patchpartititle
  {{\fontsize{19pt}{16pt}\selectfont\sffamily\bfseries\color{black}\titlelist}\par}
  {{\rmfamily\fontsize{20pt}{23pt}\selectfont
      \bfseries\color{black}\titlelist\par}%
   \ifdefempty{\titlesubtitlelist}{}{%
     \vskip 0.10cm
     {\rmfamily\fontsize{12pt}{15pt}\selectfont
       \mdseries\color{ifmmedgray}\titlesubtitlelist\par}%
   }}
\patchpartititle{\vspace{0.15cm}}{\vspace{0.10cm}}
\patchpartititle{\vspace{0.5cm}}{\vspace{0.45cm}}
\patchpartititle{\vskip 0.3cm}{\vskip 0.36cm}
\patchpartititle{\vskip 0.15cm}{\vskip 0.16cm}
\patchpartititle{\vskip 0.05cm}{\vskip 0.04cm}
\patchpartititle{\vskip 0.25cm}{\vskip 0.28cm}
\patchpartititle{\vskip 0.3cm}{\vskip 0.40cm}

\usetikzlibrary{arrows.meta,calc,decorations.pathreplacing}
\definecolor{ink}{HTML}{111111}
\definecolor{fill1}{HTML}{E4E4E4}
\definecolor{muted}{HTML}{6E6E6E}
\tikzset{
  st/.style={draw=ink,line width=0.9pt,rounded corners=3pt,minimum width=1.3cm,minimum height=0.8cm,
    fill=white,align=center,inner sep=2pt,font=\fontsize{12}{14}\selectfont},
  done/.style={st,fill=fill1},
  trained/.style={st,line width=1.6pt},
  add/.style={circle,draw=ink,line width=0.9pt,inner sep=0pt,minimum size=0.46cm,fill=white,
    font=\fontsize{12}{12}\selectfont},
  arr/.style={-{Stealth[length=6pt,width=5pt]},line width=1pt,draw=ink},
  pair/.style={{Stealth[length=6pt,width=5pt]}-{Stealth[length=6pt,width=5pt]},line width=1pt,draw=ink,dashed},
  mod/.style={font=\fontsize{12}{14}\selectfont\itshape,text=ink,fill=white,inner sep=1.5pt},
  note/.style={font=\fontsize{11.5}{13}\selectfont,text=muted,align=center},
  emb/.style={font=\fontsize{12}{14}\selectfont,text=ink},
  title/.style={font=\fontsize{13.5}{16}\selectfont\bfseries,anchor=west},
  tier/.style={note,rotate=90},
  formula/.style={font=\fontsize{11.5}{14}\selectfont,text=ink,anchor=north},
  call/.style={draw=muted,line width=0.9pt,dashed,rounded corners=6pt},
  loss/.style={emb,anchor=south},
  ulabel/.style={emb,inner sep=1pt,anchor=east},
  rerun/.style={st,double=white,double distance=1.2pt},
  plain/.style={st},
  mtpbox/.style={st,draw=mtp},
  brace/.style={decorate,decoration={brace,mirror,amplitude=5pt},draw=muted,line width=0.8pt},
}
\newcommand{\ulab}[2]{\node[ulabel] at ($(#1.north)+(-0.1,0.62)$) {$#2$};}
\newcommand{\sg}[1]{\fill[white] (#1) circle (0.15); \draw[ink,line width=0.9pt] (#1) circle (0.15);
  \draw[ink,line width=0.9pt] ($(#1)+(-0.1,-0.1)$) -- ($(#1)+(0.1,0.1)$);}

\definecolor{trunk}{HTML}{1F5FA8}
\definecolor{mtp}{HTML}{C0661A}
\definecolor{pmap}{HTML}{2E8B3E}
\definecolor{fuse}{HTML}{7A3FA0}
\definecolor{sg}{HTML}{8A8A8A}
\newcommand{\T}[1]{#1}
\newcommand{\M}[1]{\leavevmode{\color{mtp}#1}}
\newcommand{\Pm}[1]{#1}
\newcommand{\F}[1]{#1}
\newcommand{\SG}[1]{{\color{sg}\operatorname{sg}(#1)}}
\newcommand{\rms}{\operatorname{RMSNorm}}

\setlist{itemsep=1pt,topsep=2pt,leftmargin=*}
\title{Feedback Shortcut: Speculative Decoding for Latent Feedback Transformers}
\metadata[\faGithub\ Code]{\url{https://github.com/ifm-ai/LoRM}}
\date{November 20, 2026\qquad{\sffamily\bfseries\color{black}\faEnvelope[regular]\ Correspondence:} \email{huskydogewoof@gmail.com}}
\hypersetup{pdftitle={Feedback Shortcut: Speculative Decoding for Latent Feedback Transformers}, pdfauthor={Benhao Huang, Xuezhe Ma, Hector Liu}}
\abstract{
Feeding each position's top-layer hidden state back into the next position's input improves Transformer performance. However, this feedback breaks speculative decoding, since verifying a drafted token requires the top-layer state from its preceding position, which forces the draft to rely on an approximate stand-in. Training the model to adapt to this stand-in can lower the exact-decoding performance ceiling, while using it at inference time further introduces approximation error. We propose \textbf{Feedback Shortcut} to address both bottlenecks. First, \textbf{State alignment} trains a lightweight projection from multi-token prediction head states to stand-ins, leaving base model training untouched. Second, \textbf{Pipelined correction} re-evaluates each draft using its predecessor's newly computed state—a much closer proxy to true feedback. Running this correction inside the verification pass of subsequent drafts improves accuracy with zero extra calls. Across 135M and 780M scales, Feedback Shortcut stays within $10^{-4}$ bits per byte of the exact-decoding ceiling on Paloma and accelerates decoding by $2.4\times$–$2.8\times$ for only $\sim 2\%$ extra training compute. At large batch sizes, skipping correction still preserves Transformer speed and over $80\%$ of latent feedback's gains.
}
\author[\(\ddagger\ast\)\,{\faEnvelope[regular]}]{\href{https://huskydoge.github.io/}{Benhao Huang}}\authorsep{, }
\author[\(\ddagger\dagger\)]{Xuezhe Ma}\authorsep{, }
\author[\(\ddagger\)]{Hector Liu}\authorsep{}
\affiliation[\(\ddagger\)]{Institute of Foundation Models}\affiliationsep{, }
\affiliation[\(\dagger\)]{USC}\affiliationsep{, }
\affiliation[\(\ast\)]{CMU}

\begin{document}
\maketitle

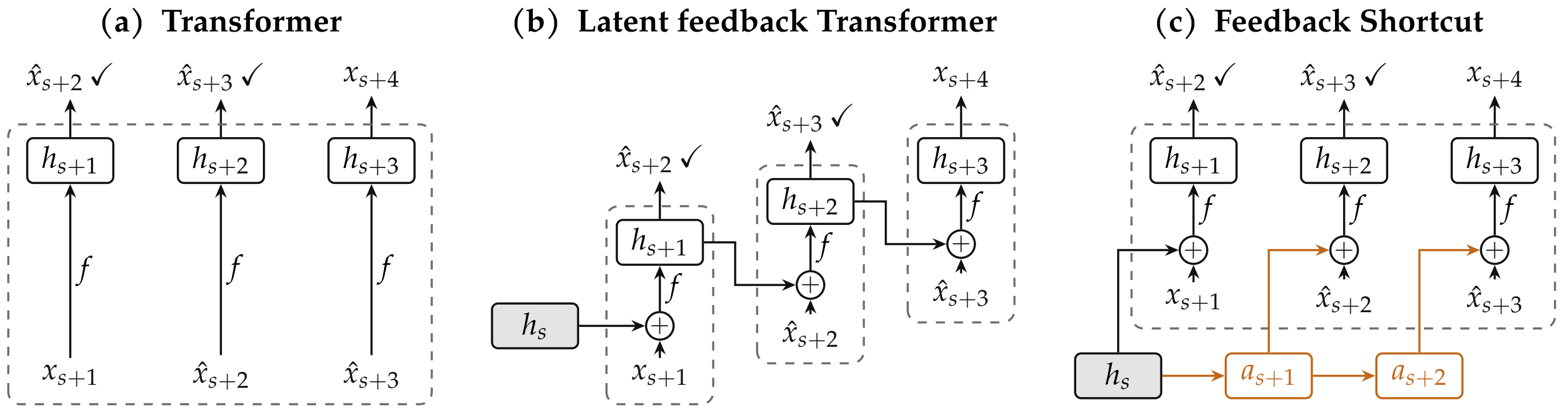
\begin{figure}[H]
\centering
\resizebox{\linewidth}{!}{% Teaser: verifying the drafts of positions s+2 and s+3 after the committed token of s+1. A position is a column and a
% dashed box one forward call: one call for a Transformer, one call per position with latent feedback (each column
% starts higher than the one it waits for), one call again once drafted positions read stand-ins. Panels share one
% height, so the one-call columns are drawn taller. Above each column is what its top state decides: the next draft
% checked, or the next token after the last draft. Drawn near print size.
\begin{tikzpicture}[
  tst/.style={draw=ink,line width=0.7pt,rounded corners=2pt,minimum width=1.0cm,minimum height=0.52cm,fill=white,
    inner sep=1pt,font=\fontsize{10}{12}\selectfont},
  tdone/.style={tst,fill=fill1},
  tmtp/.style={tst,draw=mtp,text=mtp},
  tmod/.style={inner sep=1.5pt,font=\fontsize{10}{12}\selectfont},
  tadd/.style={circle,draw=ink,line width=0.7pt,inner sep=0pt,minimum size=0.32cm,fill=white,
    font=\fontsize{8}{8}\selectfont},
  tarr/.style={-{Stealth[length=4.5pt,width=3.8pt]},line width=0.7pt,draw=ink},
  ttok/.style={font=\fontsize{10}{12}\selectfont,text=ink,inner sep=2pt},
  ttitle/.style={font=\fontsize{10.5}{13}\selectfont\bfseries,anchor=base},
  tcall/.style={draw=muted,line width=0.7pt,dashed,rounded corners=4pt}]
% \vcol{k}{y base}{y of h above the base}{1 with the feedback sum}
\newcommand{\vcol}[4]{%
  \pgfmathsetmacro{\x}{1.75*#1}
  \node[ttok] (t#1) at (\x,#2) {$\ifnum#1=1 x\else\hat x\fi_{s+#1}$};
  \node[tst] (h#1) at (\x,#2+#3) {$h_{s+#1}$};
  \ifnum#4=1 \node[tadd] (p#1) at (\x,#2+0.58) {$+$}; \draw[tarr] (t#1) -- (p#1);
    \draw[tarr] (p#1) -- node[tmod,right] {$f$} (h#1);
  \else \draw[tarr] (t#1) -- node[tmod,right] {$f$} (h#1); \fi}
% \vouts: each top state checks the draft of the next position; the last one gives the next token.
\newcommand{\vouts}{\foreach \k/\o in {1/{\hat x_{s+2}\,\checkmark},2/{\hat x_{s+3}\,\checkmark},3/{x_{s+4}}} {
    \node[ttok] (o\k) at ($(h\k.north)+(0,0.69)$) {$\o$}; \draw[tarr] (h\k) -- (o\k);}}
\begin{scope}
  \foreach \k in {1,2,3} {\vcol{\k}{0}{2.5}{0}}
  \vouts
  \draw[tcall] (1.03,-0.33) rectangle (5.95,2.92);
  \node[ttitle] at (3.49,4.0) {(a)\ \ Transformer};
\end{scope}
\begin{scope}[shift={(6.85,0)}]
  \foreach \k/\y in {1/0,2/0.475,3/0.95} {\vcol{\k}{\y}{1.55}{1}
    \draw[tcall] (1.75*\k-0.62,\y-0.33) rectangle (1.75*\k+0.62,\y+1.97);}
  \vouts
  \node[tdone] (r1) at (0.3,0.58) {$h_s$}; \draw[tarr] (r1.east) -- (p1.west);
  \draw[tarr] (h1.east) -| (2.625,1.055) -- (p2.west); \draw[tarr] (h2.east) -| (4.375,1.53) -- (p3.west);
  \node[ttitle] at (2.84,4.0) {(b)\ \ Latent feedback Transformer};
\end{scope}
\begin{scope}[shift={(13.05,0.88)}]
  \foreach \k in {1,2,3} {\vcol{\k}{0}{1.62}{1}}
  \vouts
  \draw[tcall] (1.03,-0.33) rectangle (5.95,2.04);
  \node[tdone] (r1) at (0.875,-0.88) {$h_s$}; \draw[tarr] (r1.north) |- (p1.west);
  \node[tmtp] (r2) at (2.625,-0.88) {$a_{s+1}$}; \draw[tarr,draw=mtp] (r2.north) |- (p2.west);
  \node[tmtp] (r3) at (4.375,-0.88) {$a_{s+2}$}; \draw[tarr,draw=mtp] (r3.north) |- (p3.west);
  \draw[tarr,draw=mtp] (r1) -- (r2); \draw[tarr,draw=mtp] (r2) -- (r3);
  \node[ttitle] at (3.2,3.12) {(c)\ \ Feedback Shortcut};
\end{scope}
\end{tikzpicture}}
\caption{Draft token verification in speculative decoding. $\checkmark$ marks an accepted draft, and each dashed box represents a single forward pass. (a) Standard Transformers only condition on token embeddings, allowing a single pass to verify all drafts in parallel. (b) With latent feedback, each position depends on the preceding position's top-layer state, requiring a separate pass per draft. (c) \textbf{Feedback Shortcut} supplies drafted positions with stand-ins $a_{s+k}$ projected from MTP heads so one pass verifies both drafts, while subsequent passes refine them concurrently with verifying incoming drafts in the way of pipelined correction (Section~\ref{sec:pipeline}).}
\label{fig:teaser}
\end{figure}

\section{The problem}
\label{sec:problem}
In latent feedback Transformers~\citep{fan2020feedback,wang2026fullbandwidth,bytedance2026recycler}, the top-layer state of each position is fed to the input of the next position. They perform better than a standard Transformer of the same memory cost. At decoding time, specifically, the transformer blocks $\T{f}$ map the input $u_t$ of position $t$ to its top-layer state $h_t$. The input adds the previous state, normalized by RMSNorm and projected by $\F{W}$, to the token embedding $e_t$:
\begin{equation}
u_t = e_t + \F{W}\,\rms(h_{t-1}). \label{eq:decode}
\end{equation}
Because $u_t$ depends on $h_{t-1}$, training under Eq.~\ref{eq:decode} processes one position at a time. A sequence of length $T$ requires $T$ sequential forward calls instead of one parallel call. Current training~\citep{wang2026fullbandwidth} approximates Eq.~\ref{eq:decode} with fixed-point iteration. In each pass, we compute all positions in parallel from the previous pass's states, and the fixed point of the iteration satisfies Eq.~\ref{eq:decode}:
\begin{equation}
\Big[\,u^{(p+1)}_t = e_t + \F{W}\,\rms\big(h^{(p)}_{t-1}\big),\quad h^{(p)}_{t-1}=\T{f}\big(u^{(p)}_{t-1}\big)\,\Big] \;\xrightarrow{\;p\,\to\,\infty\;}\; u^{(*)}_t = e_t + \F{W}\,\rms\big(h^{(*)}_{t-1}\big). \label{eq:fbt}
\end{equation}

The first pass reads no feedback, $u^{(1)}_t = e_t$. In pass $p = 2, 3, \dots$, we read the top states of the previous pass, which converge to the $h_{t-1}$ of Eq.~\ref{eq:decode}. In practice, two to three passes come close to the fixed point~\citep{wang2026fullbandwidth}.

Speculative decoding~\citep{leviathan2022fast,chen2023acceleratinglargelanguagemodel} drafts $K$ tokens and verifies them in one forward call, with all positions in parallel (Figure~\ref{fig:teaser}a). But in a latent feedback Transformer, a drafted position $s{+}1{+}k$ needs $h_{s+k}$, and that state is still being computed in the same call (Figure~\ref{fig:teaser}b). So, to verify all the drafts in that call, each drafted position needs to read a \emph{stand-in} $a_{s+k}$ instead of $h_{s+k}$,
\begin{equation}
u_{s+1+k} = e(\hat x_{s+1+k}) + \F{W}\,\rms(a_{s+k}), \label{eq:verify}
\end{equation}
where $k = 1, \dots, K$. Every state computed from a stand-in deviates from exact decoding, leaving a gap.

\textbf{Ceiling and gap.} We measure quality in bits per byte (BPB), where lower is better. The \emph{ceiling} $B(\cdot)$ of a model is the BPB of its exact decoding, the best quality it can serve. Regular speculative decoding serves the ceiling. With latent feedback, stand-ins let us verify in parallel again, but at a cost in quality. Let $\theta$ be the \emph{feedback model}, the latent feedback Transformer with MTP heads trained without considering its stand-in. Let $\hat\theta$ be the model trained for its stand-in $a$, and $B_a(\hat\theta)$ the BPB that speculative decoding serves with $a$. The gap to the \emph{feedback ceiling} $B(\theta)$ splits into two parts:
\begin{equation}
\underbrace{B_a(\hat\theta) - B(\theta)}_{\text{gap}} = \underbrace{B(\hat\theta) - B(\theta)}_{\text{ceiling drop } \Delta_a} + \underbrace{B_a(\hat\theta) - B(\hat\theta)}_{\text{approximation error } \epsilon_a}. \label{eq:gap}
\end{equation}
The ceiling drop $\Delta_a$ arises in training, when the model adjusts to its stand-in. The approximation error $\epsilon_a$ arises in decoding, from reading the stand-in in drafted positions. Within one call, the stand-in's reads add a \emph{one-call error} $\epsilon^1_a$. Left uncorrected, $\epsilon_a$ may accumulate and exceed $\epsilon^1_a$, because later calls also read the states computed from stand-ins.

\textbf{Questions.} Which stand-in keeps both parts of the gap small while decoding quickly with minimal training cost (Section~\ref{sec:standins})? And how can we remove the remaining approximation error in decoding without extra calls (Section~\ref{sec:pipeline})? Our answer is \emph{Feedback Shortcut} (Figure~\ref{fig:teaser}c). We let drafted positions read a trained stand-in instead of waiting for their feedback, and we correct them in later calls in parallel.

\section{Feedback stand-ins}
\label{sec:standins}
We compare four candidates for the stand-in. They differ in how we train the state that verification reads, and in whether that training also updates the model itself, its blocks $\T{f}$ and MTP heads.

\textbf{Zeros}, $a_{s+k} = 0$. We drop the feedback in drafted positions and do not change training.

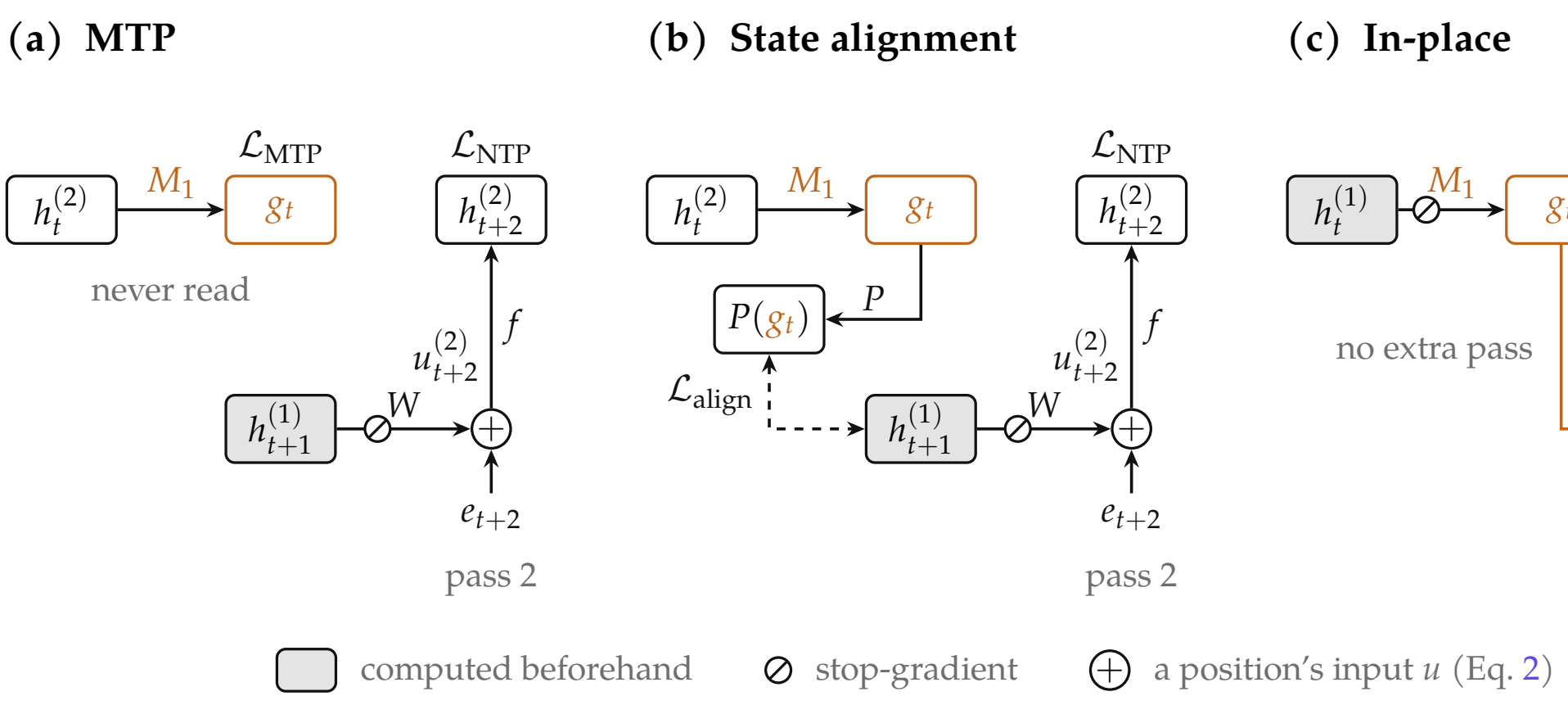
\begin{figure}[t]
\centering
\resizebox{0.85\linewidth}{!}{% The main paper's training figure: fig_paper_training.tex without the separate substitution pass, which only the
% short paper shows. Position t+2 of the pass that carries the next-token loss, the MTP chain cut to its first head.
% \reader{x}{pass label}{output}{input}: the position t+2 of a pass; the input u (Eq. 1) is the sum at the bottom.
\newcommand{\reader}[4]{
  \node[emb] (e) at (#1,-1.05) {$e_{t+2}$};
  \node[add] (a) at (#1,0) {$+$};
  \node[st] (o) at (#1,2.6) {$#3$};
  \draw[arr] (e) -- (a);
  \draw[arr] (a) -- node[mod,right=2pt] {$f$} (o);
  \node[note,anchor=north] at (#1,-1.45) {#2};
  \ulab{a}{#4}
}
\begin{tikzpicture}
% ---- (a) MTP ----
\begin{scope}
  \node[title] at (-1.5,4.6) {(a)\ \ MTP};
  \reader{4.4}{pass 2}{h^{(2)}_{t+2}}{u^{(2)}_{t+2}}
  \node[loss] at (o.north) {$\mathcal{L}_{\mathrm{NTP}}$};
  \node[done] (s) at (1.9,0) {$h^{(1)}_{t+1}$};
  \draw[arr] (s) -- node[mod,above=2pt] {$W$} (a);
  \sg{3.05,0}
  \node[st] (q) at (-0.7,2.6) {$h^{(2)}_{t}$};
  \node[st,draw=mtp] (g) at (1.9,2.6) {$\M{g_t}$};
  \draw[arr] (q) -- node[mod,above=2pt] {$\M{M_1}$} (g);
  \node[loss] at (g.north) {$\mathcal{L}_{\mathrm{MTP}}$};
  \node[note,anchor=north] at (0.6,1.95) {never read};
\end{scope}
% ---- (b) state alignment ----
\begin{scope}[shift={(7.6,0)}]
  \node[title] at (-1.5,4.6) {(b)\ \ State alignment};
  \reader{4.4}{pass 2}{h^{(2)}_{t+2}}{u^{(2)}_{t+2}}
  \node[loss] at (o.north) {$\mathcal{L}_{\mathrm{NTP}}$};
  \node[done] (s) at (1.9,0) {$h^{(1)}_{t+1}$};
  \draw[arr] (s) -- node[mod,above=2pt] {$W$} (a);
  \sg{3.05,0}
  \node[st] (q) at (-0.7,2.6) {$h^{(2)}_{t}$};
  \node[st,draw=mtp] (g) at (1.9,2.6) {$\M{g_t}$};
  \draw[arr] (q) -- node[mod,above=2pt] {$\M{M_1}$} (g);
  \node[st] (p) at (0.1,1.3) {$\Pm{P}(\M{g_t})$};
  \draw[arr] (g.south) |- node[mod,above=1pt,pos=0.75] {$\Pm{P}$} (p.east);
  \draw[pair] (p.south) -- (p.south |- s.west) -- (s.west);
  \node[emb,anchor=east] at (0.05,0.4) {$\mathcal{L}_{\mathrm{align}}$};
\end{scope}
% ---- (c) in-place substitution ----
\begin{scope}[shift={(15.2,0)}]
  \node[title] at (-1.5,4.6) {(c)\ \ In-place};
  \reader{4.4}{pass 2}{h^{(2)}_{t+2}}{u^{(2)}_{t+2}}
  \node[loss] at (o.north) {$\mathcal{L}_{\mathrm{NTP}}$};
  \node[done] (q) at (-0.7,2.6) {$h^{(1)}_{t}$};
  \node[st,draw=mtp] (g) at (1.9,2.6) {$\M{g_t}$};
  \draw[arr] (q) -- node[mod,above=2pt] {$\M{M_1}$} (g);
  \sg{0.3,2.6}
  \draw[arr,draw=mtp] (g.south) |- node[mod,above=2pt,pos=0.75] {$W$} (a.west);
  \node[note] at (0.4,0.9) {no extra pass};
\end{scope}
% ---- legend; \& because the figure is read inside \resizebox's argument ----
\matrix[anchor=north,column sep=0.15cm,ampersand replacement=\&,nodes={anchor=center},
    leg/.style={note,text height=1.7ex,text depth=0.5ex}] at (9.5,-2.4) {
  \node[done,minimum width=0.7cm,minimum height=0.5cm] {}; \& \node[leg] {computed beforehand\hspace{1.4em}}; \&
  \sg{0,0} \& \node[leg] {stop-gradient\hspace{1.4em}}; \&
  \node[add] {$+$}; \& \node[leg] {a position's input $u$ (Eq.~\ref{eq:fbt})}; \\
};
\end{tikzpicture}}
\caption{How we train the stand-in, illustrated with one MTP head. (a) With MTP alone, $\M{g_t}$ drafts a token, but it is not trained to be feedback. (b) State alignment pulls $\Pm{P}(\M{g_t})$ toward $h^{(1)}_{t+1}$, which training uses as feedback (Eq.~\ref{eq:fbt}). (c) With in-place substitution, pass 2 itself learns to read $\M{g_t}$ as the feedback through the next-token loss.}
\label{fig:training}
\end{figure}
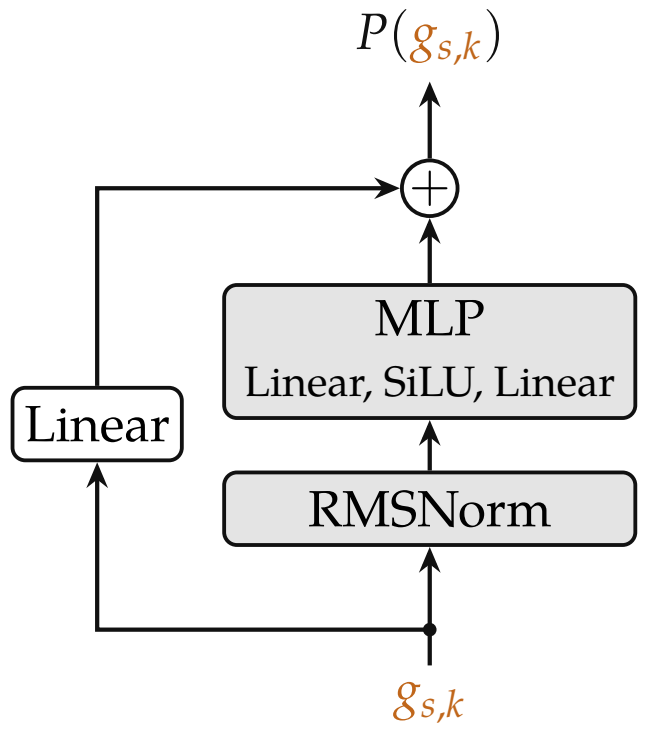
\begin{wrapfigure}{r}{0.3\linewidth}
\vspace{-12pt}
\centering
\resizebox{0.8\linewidth}{!}{% State alignment's map P (model/mtp.py FeedbackMLP) drawn bottom-to-top as LLaMA-style block diagrams draw a
% feed-forward: P(g) = W_skip g + W_down SiLU(W_up RMSNorm(g)), the RMSNorm without a learned scale; the MLP is one block
% and the skip is the Linear box on the left.
\begin{tikzpicture}[layer/.style={draw=ink,line width=0.9pt,rounded corners=3pt,minimum width=3.4cm,minimum height=0.6cm,
    inner sep=2pt,fill=fill1,font=\fontsize{12}{14}\selectfont}]
  \node[emb] (in) at (0,-0.55) {$\M{g_{s,k}}$};
  \node[layer] (norm) at (0,1.05) {RMSNorm};
  \node[layer,minimum height=1.1cm,align=center] (mlp) at (0,2.35) {MLP\\[-1pt]{\fontsize{10}{12}\selectfont Linear, SiLU, Linear}};
  \node[add] (s) at (0,3.7) {$+$};
  \node[emb] (out) at (0,4.95) {$\Pm{P}(\M{g_{s,k}})$};
  \draw[arr] (in) -- (norm); \draw[arr] (norm) -- (mlp); \draw[arr] (mlp) -- (s); \draw[arr] (s) -- (out);
  \node[layer,fill=white,minimum width=1.4cm] (skip) at (-2.75,1.75) {Linear};
  \fill[ink] (0,0.05) circle (1.6pt);
  \draw[arr] (0,0.05) -| (skip.south);
  \draw[arr] (skip.north) |- (s.west);
\end{tikzpicture}}
\caption{State alignment's map $\Pm{P}$.}
\label{fig:map}
\end{wrapfigure}
\textbf{MTP state}, $a_{s+k} = \M{g_{s,k}}$, is the state of DeepSeek-style MTP heads~\citep{gloeckle2024,deepseekai2024deepseekv3technicalreport}: $\M{g_{t,1}} = \M{M_1}(h_t, e_{t+1})$, $\M{g_{t,k}} = \M{M_k}(\M{g_{t,k-1}}, e_{t+k})$, and head $k$ drafts from $\M{g_{t,k}}$. We call the heads in sequence the \emph{chain}. The MTP loss trains $\M{g}$ to predict a token, not to be read as feedback (Figure~\ref{fig:training}a).

\textbf{State alignment}, $a_{s+k} = \Pm{P}(\M{g_{s,k}})$. We read the MTP state through a learned map $\Pm{P}$ and pull $\Pm{P}(\M{g_{t,k}})$ toward $h^{(1)}_{t+k}$ with a smooth $\ell_1$ loss, where $h^{(1)}_{t+k}$ is the first-pass state that the second pass reads (Figure~\ref{fig:training}b). $\Pm{P}$ is a linear skip plus an MLP that starts as the identity (Figure~\ref{fig:map}). This loss trains only the map, with the blocks $\T{f}$ and the heads untouched.

\textbf{In-place substitution}, $a_{s+k} = \M{g_{s,k}}$. We train the MTP state itself to be read as feedback. We divide each sequence into random-length training windows, with at most $K$ stand-ins per window. In a window anchored at $s$, position $s{+}1{+}k$ reads
\begin{equation}
u^{(2)}_{s+1+k} = e_{s+1+k} + \F{W}\,\rms\big(\M{g_{s,k}}\big), \quad \text{with} \quad \M{g_{s,1}} = \M{M_1}\big(\SG{h^{(1)}_s},\, e_{s+1}\big), \label{eq:inplace}
\end{equation}
and the next-token loss trains the heads to provide feedback (Figure~\ref{fig:training}c). We also test two milder variants, which substitute only in half the sequences of each update or only during the final learning-rate decay.

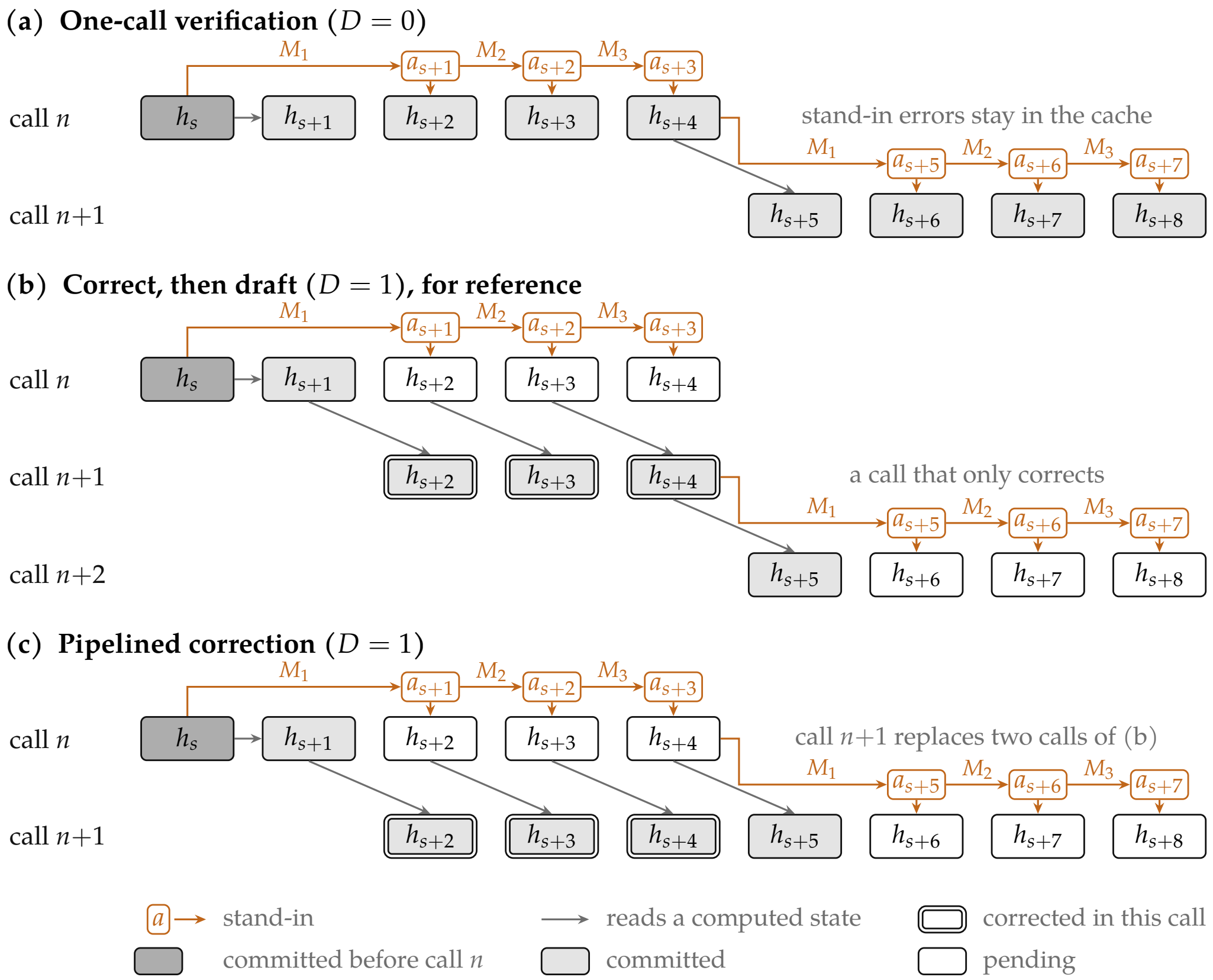
\begin{figure}[t]
\centering
\resizebox{0.78\linewidth}{!}{% The schedule of pipelined correction with K = 3: the positions s .. s+8 of Figure 1 across, one row per call, and the
% feedback each state reads drawn as an arrow into it, from the state of the position before (grey) or from a stand-in
% (orange). Against it, one-call verification and correcting before drafting again (a reference schedule). All drafts
% are accepted. Facts follow spec_fast.py merged(): with D = 1, call n commits s+1, which reads the exact h_s, and keeps
% s+2 .. s+4 pending; their re-run reads the states call n computed, which makes s+2 exact and s+3, s+4 depth 1, so
% they commit after it, together with s+5, which reads h_{s+4} from the call before.
\begin{tikzpicture}[
    big/.style={font=\fontsize{14}{17}\selectfont},
    cell/.style={st,big,minimum width=1.75cm,minimum height=0.82cm,inner sep=1pt},
    % The double border reaches 1.1pt past the path on each side: shrink the path and move the anchors out so a
    % re-run box looks as large as a plain one and arrow tips stop at its outer edge.
    rer/.style={cell,double=white,double distance=1.3pt,minimum width=1.75cm-2.2pt,minimum height=0.82cm-2.2pt,
      outer sep=1.55pt},
    com/.style={fill=fill1},
    pre/.style={fill=black!32},
    stand/.style={draw=mtp,line width=0.9pt,rounded corners=3pt,fill=white,text=mtp,big,inner sep=2.5pt,
      minimum height=0.55cm},
    readarr/.style={arr,draw=muted},
    standarr/.style={arr,draw=mtp},
    rowlab/.style={emb,big,anchor=west},
    ptitle/.style={title,font=\fontsize{15}{18}\selectfont\bfseries},
    gap/.style={note,font=\fontsize{13.5}{16}\selectfont,anchor=center}]
% \cx{k}: the x of position s+k (k = 0 is s).
\newcommand{\cx}[1]{\fpeval{3.5+2.3*(#1)}}
% \pcell{style}{k}{row y}{name}: the state of position s+k computed in the call of that row.
\newcommand{\pcell}[4]{\node[#1] (#4) at (\cx{#2},#3) {$h_{s+#2}$};}
% \stand{k}{row y}{label}: the stand-in that position s+k reads, above it, with its arrow.
\newcommand{\stand}[3]{\node[stand] (sa) at (\cx{#1},#2+0.98) {$#3$}; \draw[standarr] (sa.south) -- ++(0,-0.25);}
% \standset{source}{k}{row y}{j}{route}{reader}: the stand-ins a_{s+j} .. a_{s+j+2} that positions s+k .. s+k+2 read in place
% of h_{s+j} .. h_{s+j+2} (j = k - 1), made by the MTP heads M_1 .. M_3 from the state `source' (route: up from a
% source in the same row, or right and down from one in the row above), each with its arrow into the reader
% node {reader}<k> .. {reader}<k+2>.
\newcommand{\standset}[6]{%
  \foreach \i/\dk in {1/0,2/1,3/2} {
    \pgfmathtruncatemacro{\rep}{#4+\dk}
    \pgfmathtruncatemacro{\kk}{#2+\dk}
    \node[stand] (sa\i) at (\cx{#2+\dk},#3+0.98) {$a_{s+\rep}$};
    \draw[standarr] (sa\i.south) -- (#6\kk.north);
  }
  \ifx#5u\draw[standarr] (#1.north) |- node[mod,text=mtp,pos=0.75,above=1pt] {$M_1$} (sa1.west);
  \else\draw[standarr] (#1.east) -- ++(0.35,0) |- node[mod,text=mtp,pos=0.78,above=1pt] {$M_1$} (sa1.west);\fi
  \draw[standarr] (sa1) -- node[mod,text=mtp,above=1pt] {$M_2$} (sa2);
  \draw[standarr] (sa2) -- node[mod,text=mtp,above=1pt] {$M_3$} (sa3);}
% \reads{from}{to}: position `to' reads the state `from' computed.
\newcommand{\reads}[2]{\draw[readarr] (#1.south) -- (#2.north);}
% \exact{row y}: the state h_s, committed before call n; position s+1, whose token it predicted, reads it exactly.
\newcommand{\exact}[1]{\node[cell,pre] (hs) at (\cx{0},#1) {$h_s$};}
% ---- (a) one-call verification ----
\node[ptitle] at (-0.1,1.8) {(a)\ \ One-call verification ($D=0$)};
\node[rowlab] at (0,0) {call $n$};
\exact{0} \pcell{cell,com}{1}{0}{a1} \draw[readarr] (hs) -- (a1);
\pcell{cell,com}{2}{0}{a2}
\pcell{cell,com}{3}{0}{a3}
\pcell{cell,com}{4}{0}{a4}
\standset{hs}{2}{0}{1}{u}{a}
\node[gap] at (\cx{6.5},0) {stand-in errors stay in the cache};
\node[rowlab] at (0,-1.85) {call $n{+}1$};
\pcell{cell,com}{5}{-1.85}{b5} \reads{a4}{b5}
\pcell{cell,com}{6}{-1.85}{b6}
\pcell{cell,com}{7}{-1.85}{b7}
\pcell{cell,com}{8}{-1.85}{b8}
\standset{a4}{6}{-1.85}{5}{r}{b}
% ---- (b) correct, then draft: a reference schedule ----
\node[ptitle] at (-0.1,-3.2) {(b)\ \ Correct, then draft ($D=1$), for reference};
\node[rowlab] at (0,-4.95) {call $n$};
\exact{-4.95} \pcell{cell,com}{1}{-4.95}{c1} \draw[readarr] (hs) -- (c1);
\pcell{cell}{2}{-4.95}{c2}
\pcell{cell}{3}{-4.95}{c3}
\pcell{cell}{4}{-4.95}{c4}
\standset{hs}{2}{-4.95}{1}{u}{c}
\node[rowlab] at (0,-6.8) {call $n{+}1$};
\pcell{rer,com}{2}{-6.8}{d2} \reads{c1}{d2}
\pcell{rer,com}{3}{-6.8}{d3} \reads{c2}{d3}
\pcell{rer,com}{4}{-6.8}{d4} \reads{c3}{d4}
\node[gap] at (\cx{6.5},-6.8) {a call that only corrects};
\node[rowlab] at (0,-8.65) {call $n{+}2$};
\pcell{cell,com}{5}{-8.65}{e5} \reads{d4}{e5}
\pcell{cell}{6}{-8.65}{e6}
\pcell{cell}{7}{-8.65}{e7}
\pcell{cell}{8}{-8.65}{e8}
\standset{d4}{6}{-8.65}{5}{r}{e}
% ---- (c) pipelined correction ----
\node[ptitle] at (-0.1,-10.0) {(c)\ \ Pipelined correction ($D=1$)};
\node[rowlab] at (0,-11.75) {call $n$};
\exact{-11.75} \pcell{cell,com}{1}{-11.75}{f1} \draw[readarr] (hs) -- (f1);
\pcell{cell}{2}{-11.75}{f2}
\pcell{cell}{3}{-11.75}{f3}
\pcell{cell}{4}{-11.75}{f4}
\standset{hs}{2}{-11.75}{1}{u}{f}
\node[gap] at (\cx{6.5},-11.75) {call $n{+}1$ replaces two calls of (b)};
\node[rowlab] at (0,-13.6) {call $n{+}1$};
\pcell{rer,com}{2}{-13.6}{g2} \reads{f1}{g2}
\pcell{rer,com}{3}{-13.6}{g3} \reads{f2}{g3}
\pcell{rer,com}{4}{-13.6}{g4} \reads{f3}{g4}
\pcell{cell,com}{5}{-13.6}{g5} \reads{f4}{g5}
\pcell{cell}{6}{-13.6}{g6}
\pcell{cell}{7}{-13.6}{g7}
\pcell{cell}{8}{-13.6}{g8}
\standset{f4}{6}{-13.6}{5}{r}{g}
% ---- legend; \& because the figure is read inside \resizebox's argument ----
\matrix[anchor=north,column sep=0.18cm,ampersand replacement=\&,nodes={anchor=center},
    leg/.style={note,font=\fontsize{13.5}{16}\selectfont,text height=1.8ex,text depth=0.5ex,anchor=west},
    key/.style={minimum width=0.9cm,minimum height=0.5cm},row sep=0.12cm,column 2/.style={anchor=base west},column 4/.style={anchor=base west},column 6/.style={anchor=base west}] at (\cx{4},-14.7) {
  \node[stand] {$a$}; \draw[standarr] (0.3,0) -- (0.9,0); \& \node[leg] {stand-in\hspace{1.6em}}; \&
  \draw[readarr] (-0.45,0) -- (0.45,0); \& \node[leg] {reads a computed state\hspace{1.6em}}; \&
  \node[rer,key,minimum width=0.9cm-2.2pt,minimum height=0.5cm-2.2pt] {}; \& \node[leg] {corrected in this call}; \\
  \node[cell,pre,key] {}; \& \node[leg] {committed before call $n$\hspace{1.6em}}; \&
  \node[cell,com,key] {}; \& \node[leg] {committed\hspace{1.6em}}; \&
  \node[cell,key] {}; \& \node[leg] {pending}; \\
};
\end{tikzpicture}}
\caption{Correcting drafted positions with $K=3$ MTP heads when every draft is accepted. Each box is a state computed in that call. (a) One-call verification commits states that read stand-ins, so their errors stay in the cache. (b) Correcting first costs an extra call. (c) Pipelined correction saves that call by correcting while it verifies the next drafts. In (c), $h_{s+5}$ reads $h_{s+4}$ from call $n$ rather than a stand-in, so it commits with the corrected states.}
\label{fig:pipeline}
\end{figure}

\begin{table}[t]
\centering\footnotesize\setlength{\tabcolsep}{3.2pt}
\caption{Performance of Feedback Shortcut (state alignment, $D=1$) at batch size 1. We compare it with the standard Transformer, which speculates with its own MTP heads, and with the ceiling, exact decoding of the latent feedback Transformer. FLOPs is the training compute relative to the Transformer, and Feedback Shortcut's BPB is what speculative decoding serves (Appendices~\ref{app:eval} and~\ref{app:served}).}
\label{tab:main}
\begin{tabular}{lcccccc}
\toprule
 & & \multicolumn{4}{c}{BPB $\downarrow$} & \\
\cmidrule(lr){3-6}
Model & FLOPs $\downarrow$ & Paloma & Math & Code & QA & Speedup $\uparrow$ \\
\midrule
\multicolumn{7}{l}{\textit{135M}} \\
Transformer & $1.00\times$ & $1.0065{\scriptstyle\pm0.0005}$ & $0.8997{\scriptstyle\pm0.0023}$ & $0.6787{\scriptstyle\pm0.0031}$ & $1.2733{\scriptstyle\pm0.0086}$ & $2.49{\scriptstyle\pm<0.01}\times$ \\
Latent feedback, exact & $1.10\times$ & $0.9950{\scriptstyle\pm0.0005}$ & $0.8834{\scriptstyle\pm0.0017}$ & $0.6666{\scriptstyle\pm0.0024}$ & $1.2545{\scriptstyle\pm0.0106}$ & $1.00\times$ \\
Feedback Shortcut (ours) & $1.13\times$ & $0.9951{\scriptstyle\pm0.0003}$ & $0.8844{\scriptstyle\pm0.0026}$ & $0.6660{\scriptstyle\pm0.0018}$ & $1.2542{\scriptstyle\pm0.0049}$ & $2.36{\scriptstyle\pm0.02}\times$ \\
\midrule
\multicolumn{7}{l}{\textit{780M}} \\
Transformer & $1.00\times$ & $0.8379{\scriptstyle\pm0.0002}$ & $0.6678{\scriptstyle\pm0.0024}$ & $0.4690{\scriptstyle\pm0.0029}$ & $0.9639{\scriptstyle\pm0.0054}$ & $2.93{\scriptstyle\pm0.01}\times$ \\
Latent feedback, exact & $1.17\times$ & $0.8288{\scriptstyle\pm0.0003}$ & $0.6551{\scriptstyle\pm0.0020}$ & $0.4596{\scriptstyle\pm0.0005}$ & $0.9532{\scriptstyle\pm0.0049}$ & $1.00\times$ \\
Feedback Shortcut (ours) & $1.19\times$ & $0.8287{\scriptstyle\pm0.0001}$ & $0.6551{\scriptstyle\pm0.0017}$ & $0.4625{\scriptstyle\pm0.0014}$ & $0.9541{\scriptstyle\pm0.0059}$ & $2.75{\scriptstyle\pm0.04}\times$ \\
\bottomrule
\end{tabular}
\end{table}

\section{Pipelined correction}
\label{sec:pipeline}
No stand-in is perfect, so verification still deviates from exact decoding. But a stand-in is a good starting point for the iteration of Eq.~\ref{eq:fbt}. One more forward step \emph{corrects} a drafted position by reading the state that the last call computed for the position before it,
\begin{equation}
u^{[n]}_j = e(x_j) + \F{W}\,\rms\big(h^{[n-1]}_{j-1}\big), \label{eq:rerun}
\end{equation}
and repeated corrections approach exact decoding. Let $D$ be the number of corrections a position needs before we commit its prediction. If we run each correction as its own call, we need $D{+}1$ calls per draft (Figure~\ref{fig:pipeline}b). Instead, \emph{pipelined correction} corrects inside the next verification call. It corrects the positions still pending from the previous call and verifies the new drafts in the same call (Figure~\ref{fig:pipeline}c).

Let $d$ count the corrections a state has received so far, with $d = 0$ for a state that reads a stand-in. In a call, we commit a position's prediction only if it passes two checks. The drafted token must equal the token we commit there, and the state must have been corrected at least $D$ times. For call $n$ and position $j$,
\begin{equation}
x_{j+1} = \arg\max_v\, p\big(v \mid h^{[n]}_j\big) \quad \text{if} \quad \hat x_j = x_j \;\text{ and }\; d^{[n]}_j \ge D, \qquad d^{[n]}_j = d^{[n-1]}_{j-1} + 1. \label{eq:commit}
\end{equation}
If a draft fails the first check, we reject it and all later drafts. As in standard speculative decoding, the prediction at the position before it takes its place, and drafting resumes after it in the next call. Each call still commits at least one token. $D = 0$ is verification without corrections, and $D = \infty$ is exact decoding. Algorithm~\ref{alg:pipeline} in Appendix~\ref{app:eval} gives the full loop.

Additionally, pipelined correction only maintains a sliding window of at most $(K{+}1)(D{+}1)$ pending positions, so the memory and batch size barely change. Each call recomputes these pending positions along with the new drafts, up to $(K{+}1)(D{+}1)$ positions. These extra positions add little latency when reading a long KV cache dominates~\citep{sadhukhan2025magicdec}, especially in agentic workloads~\citep{copilot2026traces}.

\section{Experiments and results}
\label{sec:experiments}
\subsection{Setup}
\label{sec:setup}
\textbf{Models.} We train latent feedback Transformers of 135M and 780M parameters with $K=3$ MTP heads. The baseline is a standard Transformer with the same backbone and heads but without feedback. Appendix~\ref{app:eval} describes the architecture and training.

\textbf{Training and data.} Each run trains for one epoch on a prefix of the Dolma~3 mix~\citep{olmo2025olmo3}, 13.4B tokens for 135M and 77.8B tokens for 780M. Every setting uses the learning rate and weight decay tuned for the standard Transformer, and trains with three seeds.

% I think we can leave these details behind
% We keep the mix as released, without deduplication or decontamination. Its repeated documents and any overlap with Paloma are thus shared by every arm and do not change the comparisons between arms.

\textbf{Evaluation.} We evaluate quality on seven subsets of Paloma~\citep{magnusson2023paloma} and average the BPB over these subsets. We measure speed as the steady-state throughput of a batched engine that serves the first turns of the Spec-Bench and SPEED-Bench prompts~\citep{xia2024unlocking,abramovich2026speedbench} with continuous batching, relative to exact decoding in the same engine at the same batch size. For downstream quality, we measure the BPB of the gold answers in the Olmo~3 Base Easy suites~\citep{olmo2025olmo3}. All models decode greedily in float16. We report batch size 1 in Section~\ref{sec:results} and other batch sizes in Section~\ref{sec:ablations}. We compare exact decoding with verifying $K=3$ drafts under pipelined correction with $D = 0$ to 3 corrections. Unless noted, we report means over three seeds, with standard deviations as $\pm$ terms or error bars. Appendix~\ref{app:eval} describes the full protocol.

\begin{table}[t]
\centering\footnotesize\setlength{\tabcolsep}{3.2pt}
\caption{One-call error $\epsilon^1_a$ (Eq.~\ref{eq:gap}), in $10^{-3}$ BPB, and exact decoding BPB on Paloma. Below the Transformer, each row adds a stand-in to the feedback model, with its extra training FLOPs. For the 780M MTP state, the error is a range over seeds. Bold marks the lowest value per column.}
\label{tab:quality}
\begin{tabular}{lcccccc}
\toprule
 & \multicolumn{3}{c}{135M, 13.4B tokens} & \multicolumn{3}{c}{780M, 77.8B tokens} \\
\cmidrule(lr){2-4}\cmidrule(lr){5-7}
Setting & FLOPs $\downarrow$ & $\epsilon^1_a$ $\downarrow$ & Exact BPB $\downarrow$ & FLOPs $\downarrow$ & $\epsilon^1_a$ $\downarrow$ & Exact BPB $\downarrow$ \\
\midrule
Transformer & --- & --- & $1.0065{\scriptstyle\pm0.0005}$ & --- & --- & $0.8379{\scriptstyle\pm0.0002}$ \\
\midrule
MTP state & 0\% & $6.9{\scriptstyle\pm2.4}$ & $\mathbf{0.9950}{\scriptstyle\pm0.0005}$ & 0\% & $15$ to $1512$\textsuperscript{\ref{fn:mtp}} & $0.8288{\scriptstyle\pm0.0003}$ \\
zeros & 0\% & $10.0{\scriptstyle\pm0.5}$ & $\mathbf{0.9950}{\scriptstyle\pm0.0005}$ & 0\% & $6.8{\scriptstyle\pm<0.1}$ & $0.8288{\scriptstyle\pm0.0003}$ \\
state alignment & 2.4\% & $1.7{\scriptstyle\pm<0.1}$ & $0.9951{\scriptstyle\pm0.0003}$ & 2.0\% & $1.7{\scriptstyle\pm0.1}$ & $\mathbf{0.8287}{\scriptstyle\pm0.0001}$ \\
in-place & 8.0\% & $\mathbf{0.6}{\scriptstyle\pm0.1}$ & $0.9975{\scriptstyle\pm0.0005}$ & 6.1\% & $\mathbf{0.6}{\scriptstyle\pm<0.1}$ & $0.8306{\scriptstyle\pm0.0001}$ \\
\quad in half the sequences & 8.0\% & $1.1{\scriptstyle\pm0.1}$ & $0.9965{\scriptstyle\pm0.0003}$ & 6.1\% & $0.9{\scriptstyle\pm0.2}$ & $0.8299{\scriptstyle\pm0.0002}$ \\
\quad only during decay & 1.6\% & $0.8{\scriptstyle\pm0.1}$ & $0.9970{\scriptstyle\pm0.0006}$ & 1.2\% & $0.7{\scriptstyle\pm<0.1}$ & $0.8303{\scriptstyle\pm0.0002}$ \\
\bottomrule
\end{tabular}
\end{table}

% fig_results_quality_speed.pdf: served BPB against the batched engine's speedup at B = 1 and 256, 780M; drawn by
% proposal-results/fig_quality_speed.py outside this project.
\begin{figure}[t]
\centering
\includegraphics[width=\linewidth]{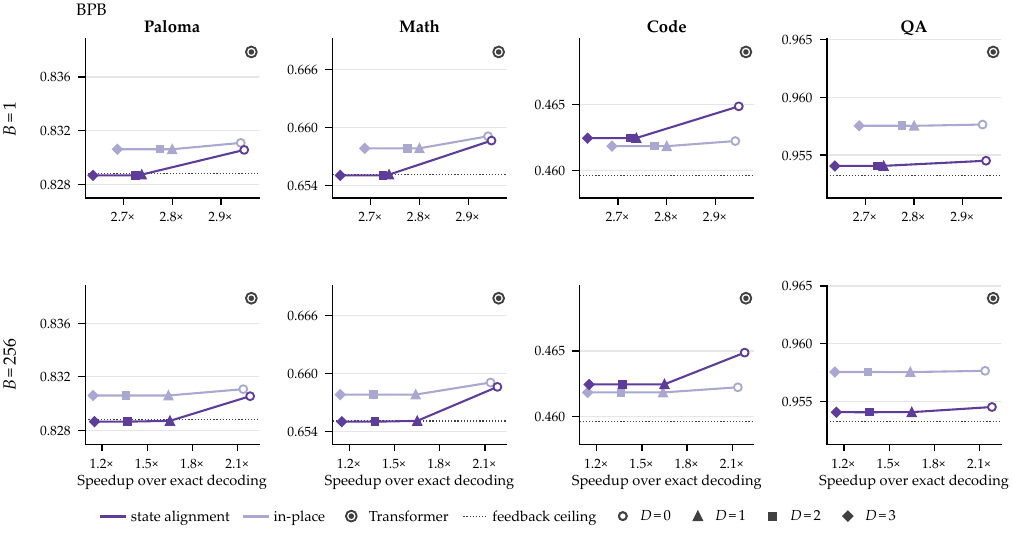}
\caption{BPB of the 780M model on each evaluation set against the speedup over exact decoding. Figure~\ref{fig:frontier135} in Appendix~\ref{app:served} shows 135M.}
\label{fig:frontier}
\end{figure}

\subsection{Feedback Shortcut against the Transformer}
\label{sec:results}
Feedback Shortcut achieves the quality of latent feedback at nearly the speculative-decoding speed of the standard Transformer. It stays close to the ceiling at both sizes ($10^{-4}$ BPB on Paloma, Table~\ref{tab:main}), keeping all of feedback's advantage over the Transformer. On downstream tasks, it also improves over the standard Transformer (Appendix~\ref{app:downstream}).

This quality costs little speed. At batch size 1, it decodes 2.36 and 2.75 times faster than exact decoding at 135M and 780M, respectively, close to the Transformer's 2.49 and 2.93 times. Training costs 1.13 and 1.19 times the Transformer's FLOPs, mostly from latent feedback's increased training costs, with state alignment adding only about 2\%.

\subsection{Ablation study}
\label{sec:ablations}
\textbf{Stand-ins.} We first ask why state alignment is the right stand-in. Within one call, the stand-ins trade one-call error against ceiling drop (Table~\ref{tab:quality}). On one hand, zeros and the MTP state as stand-ins keep the ceiling but have several times the one-call error of state alignment.\footnote{\label{fn:mtp}The 780M MTP state's error in Table~\ref{tab:quality} spans a wide range because each seed's heads set one coordinate of their state differently. The output head ignores this coordinate, so the MTP loss leaves it free, but the feedback reads it (Appendix~\ref{app:acceptance}).}  On the other hand, in-place substitution has the lowest one-call error but lowers the ceiling. 

State alignment avoids both problems. Its loss trains only the map and leaves the backbone untouched, so it keeps the ceiling with a small one-call error and extra compute. With one correction, it serves the feedback ceiling nearly as fast as in-place substitution and at a lower BPB (Figure~\ref{fig:frontier}). On downstream tasks, state alignment also keeps most of feedback's advantage (Appendix~\ref{app:downstream}).

\textbf{Number of corrections.} We next look for the number of corrections that balances quality and speed. Without correction ($D=0$), the approximation error is still nonnegligible. Fortunately, one pipelined correction ($D=1$) removes almost all of it while keeping most of the speedup (Figure~\ref{fig:frontier}).

This correction is inexpensive since it runs in the same call that verifies the next drafts, so each round needs one call instead of $D{+}1$ (Figure~\ref{fig:pipeline}). This speeds up the decoding considerably. However, it still lags behind standard speculative decoding for two reasons (Table~\ref{tab:speed}), which together trade speed for quality:
\begin{itemize}[parsep=0pt]
\item \emph{More positions in a single call.} Each call verifies the new drafts and the positions that still need correction. At small batch sizes, these positions are cheap, but they take time once each call becomes compute-bound.
\item \emph{More accurate verification.} When we check each draft against a state that reads corrected feedback instead of a stand-in, we reject some drafts that one-call verification would accept, which reduces the tokens per call from $D=0$ to $D=1$ by 2 to 4\% for the trained stand-ins and by 8 to 11\% for zeros, which simply drop the feedback at drafted positions.
\end{itemize}

To check whether errors compound once the model generates on its own, we sample free generations of 1{,}024 tokens at temperature 1 and compare the distribution of each served token with that of exact decoding (Table~\ref{tab:long} in Appendix~\ref{app:details}). With one correction, the KL divergence of state alignment is below $6\times10^{-4}$ nats per token, and a second correction lowers it about twentyfold more. So, the error remains small even for long generations. Moreover, when greedy decoding departs from exact decoding, it mostly first departs at a near-tie, a token that exact decoding itself scores within 0.1 nats of its own choice, although near-ties make up under 1\% of the positions (Table~\ref{tab:neartie} in Appendix~\ref{app:details}). So, with one correction, the free generation remains close to exact decoding, which makes $D=1$ the setting we recommend.

\begin{figure}[t]
\centering
\includegraphics[width=\linewidth]{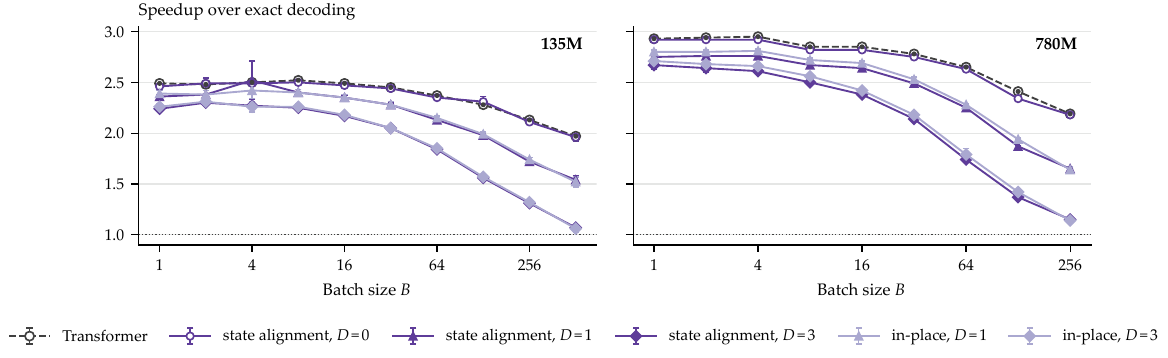}
\caption{Speedup over exact decoding against batch size $B$, excluding prefill and the CPU work between calls. One-call verification ($D=0$) is as fast as the Transformer at every batch size, and larger $D$ trades speed for accuracy.}
\label{fig:batched}
\end{figure}

\textbf{Batch size and serving.} When batching many requests, the Feedback Shortcut remains faster than exact decoding at each batch size (Figure~\ref{fig:batched}). At small ones, state alignment with one correction decodes 2.36 to 2.75 times faster than exact decoding, close to the Transformer's 2.49 to 2.93 times. However, as the batch fills the GPU, each call becomes compute-bound, and the positions that correction verifies start to take time. At batch size 256, $D=1$ is still about 1.7 times faster than exact decoding, but it lags further behind the Transformer. In this regime, a server could switch to one-call verification ($D=0$), which decodes about as fast as the Transformer but with better performance, though It keeps only 80 to 85\% of feedback's advantage (Figure~\ref{fig:frontier}). This offers heavily loaded services a way to trade model quality for serving cost.

\section{Related work}
\label{sec:related}
\textbf{Feedback Transformers.} Feedback Transformers feed the high-level state of each position to later positions and report performance gains~\citep{fan2020feedback,hao2024coconut,wang2026fullbandwidth,bytedance2026recycler,huang2026amplifier}. However, this sequential dependence breaks speculative decoding, whose verifier computes all drafted positions in a single call while each of them needs the top-layer state of the previous position. Feedback Shortcut is a step toward removing this obstacle, allowing these models to be served with speculative decoding.

\textbf{Speculative decoding.} Speculative decoding drafts tokens cheaply and verifies them with a single forward pass of the target, which preserves the target's output~\citep{stern2018blockwise,leviathan2022fast,chen2023acceleratinglargelanguagemodel}. Drafts come from small models~\citep{miao2024specinfer}, from the target's own layers~\citep{zhang2023draftverify,elhoushi2024layerskip}, from block diffusion~\citep{chen2026dflash}, or from heads on the target's states~\citep{cai2024medusa,li2024eagle,li2025eagle3}. Multi-token prediction (MTP) trains these heads along with the model, in parallel~\citep{gloeckle2024} or as a chain~\citep{deepseekai2024deepseekv3technicalreport,teoh2026multitokenpredictionright}. NextLat~\citep{teoh2026nextlatentpredictiontransformerslearn} regresses the model's next top-layer state as an auxiliary loss that shapes the model and as a source of drafts. We instead pull the MTP state toward $h^{(1)}_{t+k}$, the first-pass state that the second pass reads, and train only a map, so the stand-in can serve as feedback without changing the model (Section~\ref{sec:standins}).

\textbf{Pipelined inference.} If each output needs several sequential steps using a same function module, the next output can start before the current one finishes and share its batched call. Looped Transformers pipeline across depth~\citep{geiping2025parallel,cho2026loopspec,ha2026wavefront,liu2026loopahead,huang2026towards}, and video diffusion pipelines across denoising steps~\citep{kim2024fifo,ruhe2024rolling,chen2024diffusionforcing}. Our pipelined correction shares a similar spirit. Each call corrects the positions that read stand-ins while verifying the next drafts, making the verification more accurate without extra calls.

\section{Conclusions}
Latent feedback breaks speculative decoding because a drafted position needs to read a state that is still being computed in its verification call. We remove this obstacle by training the stand-in to be read as feedback and by correcting the drafted positions while verifying the next drafts. This recovers most of the speedup that speculative decoding gives a Transformer, at a quality close to that of the feedback model.

\paragraph{Limitations}
The Feedback Shortcut is still not exact. Closing the remaining gap needs a stand-in that adds less error while keeping the ceiling. We leave such stand-in and larger-scale experiments to future work.

% \section*{Acknowledgments}
% We thank \textcolor{sg}{\textit{[names]}} for discussions and feedback on this work. Compute was provided by \textcolor{sg}{\textit{[compute provider]}}, and this work was supported by \textcolor{sg}{\textit{[funding]}}.

\vspace{-2pt}
{\small
\bibliographystyle{unsrtnat}
\setlength{\bibsep}{1pt}
\bibliography{ref,extra}
}

\clearpage
\appendix
\section{Training and evaluation details}
\label{app:eval}
\textbf{Models.} We use 135M models with 12 blocks of 768 width each and 780M models with 24 blocks of 1536 width each. The backbone is a pre-norm decoder with RMSNorm~\citep{zhang2019rmsnorm}, rotary positions~\citep{su2021roformer}, QK-norm~\citep{henry2020qknorm} and SwiGLU~\citep{shazeer2020glu} with untied embeddings. Feedback is fed into the first block of the decoder using the linear fusion $\F{W}$ of Eq.~\ref{eq:decode}. One training pass is run with probability 0.31 and two with probability 0.69. For updates with two training passes, the first pass is run without gradient and the losses are computed in the second pass. The MTP heads are trained with loss weight 0.3 throughout training, as in DeepSeek-V4~\citep{deepseekai2026deepseekv4}.

\textbf{Training.} For training the matrices, we use NorMuon~\citep{li2025normuon}. For the embeddings and the output head, we use AdamW~\citep{loshchilov2019adamw}. We use a learning rate that increases during the first 1\% of the updates and then decays linearly to zero during the last 20\% of the updates~\citep{hagele2024scaling}. Every setting uses the learning rate and weight decay that perform best for the Transformer on a grid of one-epoch runs for each model size. These values are also the best for latent feedback at 780M, and at 135M they are 0.0025 nats per token worse on the validation set than latent feedback's best. Each update processes 256 sequences of 2{,}048 tokens. The data is a single shuffled stream of the 150B-token Dolma~3 mix, tokenized with the Llama~2 tokenizer~\citep{touvron2023llama2}.

\textbf{Stand-ins.} State alignment's map $\Pm{P}$ has 3.7M parameters at 135M and 14.9M at 780M, with hidden width 2{,}048 and 4{,}096; its RMSNorm has no learned scale. Each training window of in-place substitution extends past each of its positions with probability 0.75.

\textbf{Quality.} We score each Paloma document on its own for exact BPB, as Paloma scores documents. We measure the one-call error $\epsilon^1_a$ teacher-forced on windows of the concatenated documents, with all earlier windows being corrected, so that only the current window reads stand-ins. Downstream answers are scored as OLMES scores answers, after the document-start token the models were trained with.

\textbf{Served quality.} To measure the quality of what is served by speculative decoding, the decoding loop is run on a fixed text. For each call of the loop, $K$ tokens are generated with the MTP heads and the pending positions are verified under pipelined correction with $D$ corrections. Each committed token is equal to the corresponding token in the fixed text, and a draft is kept only while it is equal to the fixed text. Each token is scored by the distribution of its predecessor in the call that committed it. On Paloma, the loop is run on the windows of the one-call error after prefilling each window's first 33 tokens, and Figure~\ref{fig:frontier} adds the difference between served and exact BPB measured there to the exact BPB of the whole documents. For the downstream suites, the context is prefilled and the answer is decoded with the same loop, which measures the served BPB directly.

\begin{algorithm}[!htbp]
\caption{Greedy decoding with Feedback Shortcut. $\mathcal{W}$ holds the pending positions, $c$ is the last position whose state is final, and $y_j = \arg\max_v p(v \mid h_j)$.}
\label{alg:pipeline}
\begin{algorithmic}[1]
\STATE Prefill $x_{1:c}$ and commit $x_{c+1}$; $d_c \gets \infty$, $\mathcal{W} \gets \emptyset$
\REPEAT
\STATE $s \gets \max(\mathcal{W} \cup \{c\})$
\STATE $\hat x_{s+1} \gets y_s$; draft $\hat x_{s+2:s+1+K}$ and $a_{s+1:s+K}$ from $h_s$ \hfill $\triangleright$ MTP chain
\STATE $\mathcal{W} \gets \mathcal{W} \cup \{s{+}1, \dots, s{+}1{+}K\}$
\STATE $d_j \gets 0$ if $j > s{+}1$, else $d_{j-1} + 1$, for all $j \in \mathcal{W}$ \hfill $\triangleright$ Eq.~\ref{eq:commit}
\STATE Compute $h_j$ and $y_j$ for all $j \in \mathcal{W}$ in one call \hfill $\triangleright$ Eq.~\ref{eq:verify} if $d_j = 0$, else Eq.~\ref{eq:rerun}
\WHILE{$c{+}1 \in \mathcal{W}$ and $d_{c+1} \ge D$ and $\hat x_{c+1} = x_{c+1}$}
\STATE $x_{c+2} \gets y_{c+1}$; $c \gets c{+}1$ \hfill $\triangleright$ commit
\ENDWHILE
\STATE $\mathcal{W} \gets \{ j \in \mathcal{W} : j > c,\ \hat x_i = y_{i-1} \text{ for all } c < i \le j\}$ \hfill $\triangleright$ reject
\UNTIL{$x_{c+1}$ is the end token}
\end{algorithmic}
\end{algorithm}

\textbf{Decoding and timing.} All evaluations, including timings, are performed in float16, following Spec-Bench~\citep{xia2024unlocking}. The models are trained in bfloat16 mixed precision, but the largest activations are at least $7\times$ smaller than the largest value in float16, and the finer rounding of float16 keeps greedy text closer to float32 decoding than bfloat16 mixed precision. Prefill runs 4 parallel passes of Eq.~\ref{eq:fbt} over the prompt, and exact decoding uses the same prefill. Each answer is decoded to at most 256 new tokens. The single-request timings of Appendix~\ref{app:details} synchronize the GPU at the boundaries of each request, include prefill and the CPU’s bookkeeping, and count the tokens of both decoders as they are emitted.

\textbf{Batched engine.} To obtain the speeds reported in Section~\ref{sec:experiments}, we implemented a batched engine on a single GPU. It takes as input the first turns of Spec-Bench’s 480 questions as well as the 435 SPEED-Bench prompts not contained in Spec-Bench, which are randomly ordered and cycled into a queue of size $\max(300, 8B)$ requests. The engine then serves this queue of requests from $B$ slots using continuous batching. Each request is stopped at the end token or after 256 new tokens, and its slot then prefills the next prompt in the queue between calls. Each call then packs the pending positions of all slots, so each request computes only its own pending positions and drafts, and exact decoding is also run in the same engine. Throughput is the number of committed tokens per second, over the calls in which all $B$ slots are busy with requests from the queue. Each call is timed from the upload of its rows to the GPU until its summary is returned to the CPU, so prefill and the CPU’s bookkeeping between calls are not timed.

\section{Single-request timing and long generations}
\label{app:details}
Table~\ref{tab:speed} shows the decoding of 480 Spec-Bench questions, processed request by request (greedily and with sampling) and also reports the number of tokens committed per call, together with $\rho$ behind the costs of correction (see Section~\ref{sec:ablations}). Table~\ref{tab:long} shows generations of 1{,}024 tokens.

\begin{table}[!htbp]
\centering\footnotesize\setlength{\tabcolsep}{2.0pt}
\caption{Single-request decoding of Spec-Bench, timed as in Appendix~\ref{app:eval}. The bottom two blocks are greedy. For state alignment, in-place substitution and the Transformer, every Spec-Bench task lies within $0.15\times$ of the greedy speedup over all questions at $D=1$. The Transformer verifies exactly and needs no correction, and the 780M MTP state shows ranges because two of its three seeds fail as stand-ins.}
\label{tab:speed}
\begin{tabular}{lcccccccc}
\toprule
 & \multicolumn{4}{c}{135M} & \multicolumn{4}{c}{780M} \\
\cmidrule(lr){2-5}\cmidrule(lr){6-9}
Setting & $D{=}0$ & $D{=}1$ & $D{=}2$ & $D{=}3$ & $D{=}0$ & $D{=}1$ & $D{=}2$ & $D{=}3$ \\
\midrule
\multicolumn{9}{l}{\textit{Speedup, greedy $\uparrow$}} \\
Transformer & \multicolumn{4}{c}{$2.34{\scriptstyle\pm0.05}\times$} & \multicolumn{4}{c}{$2.69{\scriptstyle\pm0.01}\times$} \\
zeros & $2.26{\scriptstyle\pm0.03}\times$ & $2.07{\scriptstyle\pm0.08}\times$ & $1.99{\scriptstyle\pm0.04}\times$ & $1.93{\scriptstyle\pm0.08}\times$ & $2.51{\scriptstyle\pm0.11}\times$ & $2.29{\scriptstyle\pm0.03}\times$ & $2.22{\scriptstyle\pm0.11}\times$ & $2.15{\scriptstyle\pm0.08}\times$ \\
MTP state & $2.25{\scriptstyle\pm0.04}\times$ & $2.08{\scriptstyle\pm0.04}\times$ & $2.01{\scriptstyle\pm0.01}\times$ & $1.96{\scriptstyle\pm0.04}\times$ & $1.91$--$2.60\times$ & $1.02$--$2.39\times$ & $0.94$--$2.31\times$ & $0.93$--$2.22\times$ \\
state alignment & $2.19{\scriptstyle\pm0.06}\times$ & $2.05{\scriptstyle\pm0.06}\times$ & $2.03{\scriptstyle\pm0.02}\times$ & $1.95{\scriptstyle\pm0.07}\times$ & $2.58{\scriptstyle\pm0.02}\times$ & $2.45{\scriptstyle\pm0.02}\times$ & $2.36{\scriptstyle\pm0.01}\times$ & $2.30{\scriptstyle\pm0.03}\times$ \\
in-place & $2.21{\scriptstyle\pm0.04}\times$ & $2.11{\scriptstyle\pm0.03}\times$ & $2.02{\scriptstyle\pm0.04}\times$ & $1.99{\scriptstyle\pm0.03}\times$ & $2.54{\scriptstyle\pm0.02}\times$ & $2.40{\scriptstyle\pm0.02}\times$ & $2.28{\scriptstyle\pm0.04}\times$ & $2.28{\scriptstyle\pm0.06}\times$ \\
\quad in half the sequences & $2.20{\scriptstyle\pm0.04}\times$ & $2.12{\scriptstyle\pm0.01}\times$ & $2.09{\scriptstyle\pm0.05}\times$ & $2.01{\scriptstyle\pm0.02}\times$ & $2.56{\scriptstyle\pm0.05}\times$ & $2.45{\scriptstyle\pm0.04}\times$ & $2.40{\scriptstyle\pm0.07}\times$ & $2.26{\scriptstyle\pm0.04}\times$ \\
\quad only during decay & $2.22{\scriptstyle\pm0.02}\times$ & $2.12{\scriptstyle\pm0.05}\times$ & $1.95{\scriptstyle\pm0.06}\times$ & $1.96{\scriptstyle\pm0.01}\times$ & $2.49{\scriptstyle\pm0.03}\times$ & $2.42{\scriptstyle\pm0.01}\times$ & $2.26{\scriptstyle\pm0.01}\times$ & $2.23{\scriptstyle\pm0.05}\times$ \\
\midrule
\multicolumn{9}{l}{\textit{Speedup, sampling at temperature 1 $\uparrow$}} \\
Transformer & \multicolumn{4}{c}{$1.44{\scriptstyle\pm0.02}\times$} & \multicolumn{4}{c}{$1.87{\scriptstyle\pm0.05}\times$} \\
zeros & $1.53{\scriptstyle\pm0.02}\times$ & $1.29{\scriptstyle\pm0.02}\times$ & $1.25{\scriptstyle\pm0.02}\times$ & $1.23{\scriptstyle\pm0.03}\times$ & $1.83{\scriptstyle\pm0.03}\times$ & $1.73{\scriptstyle\pm0.01}\times$ & $1.63{\scriptstyle\pm0.01}\times$ & $1.58{\scriptstyle\pm0.05}\times$ \\
MTP state & $1.53{\scriptstyle\pm0.01}\times$ & $1.33{\scriptstyle\pm0.04}\times$ & $1.28{\scriptstyle\pm0.04}\times$ & $1.26{\scriptstyle\pm0.04}\times$ & $1.27$--$1.79\times$ & $0.88$--$1.69\times$ & $0.83$--$1.55\times$ & $0.81$--$1.48\times$ \\
state alignment & $1.51{\scriptstyle\pm0.11}\times$ & $1.37{\scriptstyle\pm0.10}\times$ & $1.32{\scriptstyle\pm0.08}\times$ & $1.32{\scriptstyle\pm0.14}\times$ & $1.83{\scriptstyle\pm0.06}\times$ & $1.75{\scriptstyle\pm0.03}\times$ & $1.70{\scriptstyle\pm0.02}\times$ & $1.66{\scriptstyle\pm0.04}\times$ \\
in-place & $1.54{\scriptstyle\pm0.14}\times$ & $1.42{\scriptstyle\pm0.12}\times$ & $1.38{\scriptstyle\pm0.12}\times$ & $1.40{\scriptstyle\pm0.12}\times$ & $1.88{\scriptstyle\pm0.04}\times$ & $1.77{\scriptstyle\pm0.03}\times$ & $1.73{\scriptstyle\pm0.02}\times$ & $1.69{\scriptstyle\pm0.03}\times$ \\
\quad in half the sequences & $1.45{\scriptstyle\pm0.03}\times$ & $1.37{\scriptstyle\pm0.02}\times$ & $1.30{\scriptstyle\pm0.02}\times$ & $1.27{\scriptstyle\pm0.03}\times$ & $1.91{\scriptstyle\pm0.05}\times$ & $1.79{\scriptstyle\pm0.05}\times$ & $1.73{\scriptstyle\pm0.05}\times$ & $1.71{\scriptstyle\pm0.05}\times$ \\
\quad only during decay & $1.53{\scriptstyle\pm0.06}\times$ & $1.42{\scriptstyle\pm0.04}\times$ & $1.36{\scriptstyle\pm0.04}\times$ & $1.36{\scriptstyle\pm0.05}\times$ & $1.77{\scriptstyle\pm0.01}\times$ & $1.62{\scriptstyle\pm0.02}\times$ & $1.58{\scriptstyle\pm0.02}\times$ & $1.56{\scriptstyle\pm0.01}\times$ \\
\midrule
\multicolumn{9}{l}{\textit{Tokens per call $\uparrow$}} \\
zeros & $3.70{\scriptstyle\pm0.04}$ & $3.41{\scriptstyle\pm0.08}$ & $3.32{\scriptstyle\pm0.09}$ & $3.27{\scriptstyle\pm0.09}$ & $3.66{\scriptstyle\pm0.01}$ & $3.27{\scriptstyle\pm0.01}$ & $3.22{\scriptstyle\pm0.02}$ & $3.19{\scriptstyle\pm0.02}$ \\
MTP state & $3.69{\scriptstyle\pm0.08}$ & $3.48{\scriptstyle\pm0.07}$ & $3.40{\scriptstyle\pm0.07}$ & $3.36{\scriptstyle\pm0.08}$ & $2.44$--$3.63$ & $1.17$--$3.28$ & $1.18$--$3.21$ & $1.17$--$3.18$ \\
state alignment & $3.71{\scriptstyle\pm0.02}$ & $3.58{\scriptstyle\pm0.04}$ & $3.52{\scriptstyle\pm0.05}$ & $3.48{\scriptstyle\pm0.05}$ & $3.64{\scriptstyle\pm0.01}$ & $3.50{\scriptstyle\pm0.04}$ & $3.45{\scriptstyle\pm0.04}$ & $3.40{\scriptstyle\pm0.05}$ \\
in-place & $3.69{\scriptstyle\pm0.04}$ & $3.58{\scriptstyle\pm0.07}$ & $3.53{\scriptstyle\pm0.08}$ & $3.49{\scriptstyle\pm0.07}$ & $3.59{\scriptstyle\pm0.01}$ & $3.51{\scriptstyle\pm0.03}$ & $3.46{\scriptstyle\pm0.02}$ & $3.43{\scriptstyle\pm0.02}$ \\
\quad in half the sequences & $3.65{\scriptstyle\pm0.07}$ & $3.58{\scriptstyle\pm0.07}$ & $3.55{\scriptstyle\pm0.07}$ & $3.51{\scriptstyle\pm0.07}$ & $3.61{\scriptstyle\pm0.03}$ & $3.52{\scriptstyle\pm0.04}$ & $3.48{\scriptstyle\pm0.04}$ & $3.44{\scriptstyle\pm0.04}$ \\
\quad only during decay & $3.69{\scriptstyle\pm0.04}$ & $3.58{\scriptstyle\pm0.03}$ & $3.54{\scriptstyle\pm0.03}$ & $3.50{\scriptstyle\pm0.02}$ & $3.60{\scriptstyle\pm0.01}$ & $3.50{\scriptstyle\pm0.02}$ & $3.45{\scriptstyle\pm0.02}$ & $3.42{\scriptstyle\pm0.02}$ \\
\midrule
\multicolumn{9}{l}{\textit{Positions verified per committed token $\rho$ $\downarrow$}} \\
zeros & $1.08$ & $1.90$ & $2.65$ & $3.38$ & $1.09$ & $1.94$ & $2.67$ & $3.39$ \\
MTP state & $1.08$ & $1.88$ & $2.63$ & $3.37$ & $1.10$--$1.64$ & $1.94$--$3.96$ & $2.67$--$4.10$ & $3.39$--$4.25$ \\
state alignment & $1.08$ & $1.85$ & $2.60$ & $3.35$ & $1.10$ & $1.87$ & $2.61$ & $3.36$ \\
in-place & $1.08$ & $1.85$ & $2.60$ & $3.35$ & $1.11$ & $1.87$ & $2.61$ & $3.35$ \\
\quad in half the sequences & $1.10$ & $1.85$ & $2.59$ & $3.34$ & $1.11$ & $1.86$ & $2.61$ & $3.35$ \\
\quad only during decay & $1.08$ & $1.85$ & $2.59$ & $3.34$ & $1.11$ & $1.87$ & $2.61$ & $3.35$ \\
\bottomrule
\end{tabular}
\end{table}

\textbf{Sampling lowers every speedup} (Table~\ref{tab:speed}). Under speculative sampling at temperature 1~\citep{leviathan2022fast,chen2023acceleratinglargelanguagemodel}, a call commits 2.3 to 2.6 tokens instead of about 3.5, so every setting decodes more slowly than under greedy decoding. At 780M, state alignment decodes within 3\% of in-place substitution's speed for every $D$; at 135M, seeds vary more than the two differ.

\begin{table}[H]
\centering\footnotesize\setlength{\tabcolsep}{5pt}
\caption{KL divergence per token from exact decoding, over 35 samples of 1{,}024 tokens at temperature 1.}
\label{tab:long}
\begin{tabular}{llcc}
\toprule
Setting & $D$ & 135M & 780M \\
\midrule
\multicolumn{4}{l}{\textit{Sampling: KL divergence from exact decoding, nats $\downarrow$}} \\
zeros & 1 & $1.8{\scriptstyle\pm<0.1}\times10^{-3}$ & $1.7{\scriptstyle\pm0.2}\times10^{-3}$ \\
state alignment & 1 & $5.0{\scriptstyle\pm0.5}\times10^{-4}$ & $5.7{\scriptstyle\pm0.3}\times10^{-4}$ \\
state alignment & 2 & $2.6{\scriptstyle\pm0.4}\times10^{-5}$ & $3.1{\scriptstyle\pm0.2}\times10^{-5}$ \\
in-place & 1 & $1.7{\scriptstyle\pm0.1}\times10^{-4}$ & $2.0{\scriptstyle\pm0.2}\times10^{-4}$ \\
in-place & 2 & $8.6{\scriptstyle\pm1.0}\times10^{-6}$ & $1.2{\scriptstyle\pm0.1}\times10^{-5}$ \\
\bottomrule
\end{tabular}
\end{table}

\textbf{Greedy decoding diverges mostly at near-ties} (Table~\ref{tab:neartie}). Near-ties are rare, accounting for under $2\%$ of positions before the first divergence and under $1\%$ after a single correction. Yet with one correction, they account for $71\%$ to $95\%$ of all first divergences—meaning a near-tie diverges hundreds to thousands of times more often than a non-tie position. Each correction pass both reduces total divergences and concentrates the remaining ones onto near-ties. Furthermore, after one correction, exact decoding assigns the served continuation a log-probability within $0.01$ nats per token of its own greedy choice from the first divergence onward. Thus, a divergence simply switches to an alternative continuation that exact decoding deems nearly equally probable.

\begin{table}[!htbp]
\centering\footnotesize\setlength{\tabcolsep}{3pt}
\caption{First departures of greedy decoding from exact decoding, over 140 continuations of 1{,}024 tokens per seed. A near-tie is a position where exact decoding's two best tokens are within 0.1 nats, and a departure counts as at a near-tie when exact decoding scores the chosen token within 0.1 nats of its own. Departed counts the continuations that depart, shares and rates count the positions up to each first departure, and rates are departures per 1{,}000 near-tie or other positions. Served and Exact are the log-probabilities per token that exact decoding gives the served continuation and its own, from the first departure on.}
\label{tab:neartie}
\begin{tabular}{llccccccc}
\toprule
Setting & $D$ & Departed $\downarrow$ & Near-ties, \% & At near-ties, \% $\uparrow$ & Near-tie rate $\downarrow$ & Other rate $\downarrow$ & Served & Exact \\
\midrule
\multicolumn{9}{l}{\textit{135M}} \\
zeros & 0 & $125{\scriptstyle\pm2}$ & $0.67{\scriptstyle\pm0.07}$ & $30{\scriptstyle\pm4}$ & $303{\scriptstyle\pm26}$ & $4.91{\scriptstyle\pm0.75}$ & $-0.057{\scriptstyle\pm0.002}$ & $-0.053{\scriptstyle\pm0.002}$ \\
zeros & 1 & $60{\scriptstyle\pm3}$ & $0.40{\scriptstyle\pm0.01}$ & $71{\scriptstyle\pm3}$ & $125{\scriptstyle\pm19}$ & $0.20{\scriptstyle\pm0.02}$ & $-0.041{\scriptstyle\pm0.004}$ & $-0.039{\scriptstyle\pm0.004}$ \\
zeros & 2 & $17{\scriptstyle\pm2}$ & $0.32{\scriptstyle\pm0.01}$ & $78{\scriptstyle\pm16}$ & $33{\scriptstyle\pm11}$ & $0.03{\scriptstyle\pm0.02}$ & $-0.035{\scriptstyle\pm0.007}$ & $-0.034{\scriptstyle\pm0.006}$ \\
state alignment & 0 & $106{\scriptstyle\pm2}$ & $0.53{\scriptstyle\pm0.08}$ & $53{\scriptstyle\pm4}$ & $284{\scriptstyle\pm37}$ & $1.30{\scriptstyle\pm0.22}$ & $-0.047{\scriptstyle\pm0.002}$ & $-0.049{\scriptstyle\pm0.001}$ \\
state alignment & 1 & $33{\scriptstyle\pm8}$ & $0.34{\scriptstyle\pm0.03}$ & $79{\scriptstyle\pm3}$ & $69{\scriptstyle\pm15}$ & $0.06{\scriptstyle\pm0.02}$ & $-0.043{\scriptstyle\pm0.006}$ & $-0.042{\scriptstyle\pm0.003}$ \\
state alignment & 2 & $10{\scriptstyle\pm3}$ & $0.31{\scriptstyle\pm0.02}$ & $90{\scriptstyle\pm2}$ & $22{\scriptstyle\pm8}$ & $0.01{\scriptstyle\pm<0.01}$ & $-0.048{\scriptstyle\pm0.012}$ & $-0.045{\scriptstyle\pm0.002}$ \\
in-place & 0 & $86{\scriptstyle\pm7}$ & $0.39{\scriptstyle\pm0.07}$ & $65{\scriptstyle\pm8}$ & $251{\scriptstyle\pm29}$ & $0.53{\scriptstyle\pm0.09}$ & $-0.053{\scriptstyle\pm0.004}$ & $-0.053{\scriptstyle\pm0.004}$ \\
in-place & 1 & $19{\scriptstyle\pm2}$ & $0.30{\scriptstyle\pm0.02}$ & $95{\scriptstyle\pm5}$ & $48{\scriptstyle\pm6}$ & $0.01{\scriptstyle\pm0.01}$ & $-0.043{\scriptstyle\pm0.007}$ & $-0.046{\scriptstyle\pm0.005}$ \\
in-place & 2 & $9{\scriptstyle\pm3}$ & $0.29{\scriptstyle\pm0.02}$ & $96{\scriptstyle\pm6}$ & $21{\scriptstyle\pm6}$ & $0.00{\scriptstyle\pm<0.01}$ & $-0.046{\scriptstyle\pm0.025}$ & $-0.041{\scriptstyle\pm0.022}$ \\
\midrule
\multicolumn{9}{l}{\textit{780M}} \\
zeros & 0 & $133{\scriptstyle\pm4}$ & $1.51{\scriptstyle\pm0.52}$ & $36{\scriptstyle\pm3}$ & $329{\scriptstyle\pm36}$ & $9.18{\scriptstyle\pm3.66}$ & $-0.063{\scriptstyle\pm0.001}$ & $-0.064{\scriptstyle\pm0.003}$ \\
zeros & 1 & $79{\scriptstyle\pm5}$ & $0.58{\scriptstyle\pm0.08}$ & $72{\scriptstyle\pm9}$ & $143{\scriptstyle\pm25}$ & $0.31{\scriptstyle\pm0.06}$ & $-0.058{\scriptstyle\pm0.004}$ & $-0.053{\scriptstyle\pm0.004}$ \\
zeros & 2 & $28{\scriptstyle\pm6}$ & $0.48{\scriptstyle\pm0.06}$ & $88{\scriptstyle\pm8}$ & $44{\scriptstyle\pm10}$ & $0.03{\scriptstyle\pm0.02}$ & $-0.054{\scriptstyle\pm0.011}$ & $-0.052{\scriptstyle\pm0.009}$ \\
state alignment & 0 & $117{\scriptstyle\pm5}$ & $0.81{\scriptstyle\pm0.16}$ & $53{\scriptstyle\pm3}$ & $286{\scriptstyle\pm36}$ & $2.04{\scriptstyle\pm0.40}$ & $-0.062{\scriptstyle\pm0.003}$ & $-0.062{\scriptstyle\pm0.003}$ \\
state alignment & 1 & $42{\scriptstyle\pm6}$ & $0.47{\scriptstyle\pm0.04}$ & $87{\scriptstyle\pm2}$ & $76{\scriptstyle\pm11}$ & $0.05{\scriptstyle\pm0.01}$ & $-0.056{\scriptstyle\pm0.008}$ & $-0.057{\scriptstyle\pm0.009}$ \\
state alignment & 2 & $15{\scriptstyle\pm5}$ & $0.44{\scriptstyle\pm0.01}$ & $90{\scriptstyle\pm5}$ & $23{\scriptstyle\pm10}$ & $0.01{\scriptstyle\pm<0.01}$ & $-0.052{\scriptstyle\pm0.006}$ & $-0.052{\scriptstyle\pm0.013}$ \\
in-place & 0 & $101{\scriptstyle\pm2}$ & $0.60{\scriptstyle\pm0.01}$ & $63{\scriptstyle\pm4}$ & $240{\scriptstyle\pm22}$ & $0.85{\scriptstyle\pm0.07}$ & $-0.062{\scriptstyle\pm0.005}$ & $-0.061{\scriptstyle\pm0.002}$ \\
in-place & 1 & $24{\scriptstyle\pm7}$ & $0.44{\scriptstyle\pm0.02}$ & $92{\scriptstyle\pm6}$ & $42{\scriptstyle\pm16}$ & $0.01{\scriptstyle\pm0.01}$ & $-0.063{\scriptstyle\pm0.012}$ & $-0.069{\scriptstyle\pm0.013}$ \\
in-place & 2 & $10{\scriptstyle\pm6}$ & $0.44{\scriptstyle\pm0.02}$ & $93{\scriptstyle\pm12}$ & $18{\scriptstyle\pm12}$ & $0.00{\scriptstyle\pm<0.01}$ & $-0.058{\scriptstyle\pm0.015}$ & $-0.056{\scriptstyle\pm0.015}$ \\
\bottomrule
\end{tabular}
\end{table}

\section{Drafting Quality Does Not Determine Readability}
\label{app:acceptance}
Table~\ref{tab:acceptance} compares each head's drafting quality against its single-call error and exact BPB across seeds. Within each model scale, every seed and setting achieves nearly identical drafting quality, yet the error spans over three orders of magnitude. Because the MTP loss constrains an MTP head's hidden state solely through the output head, the remaining dimensions of the state are left unconstrained. State alignment reads these head states via a learned projection while keeping their training unchanged; in-place substitution instead exploits this freedom to make the raw state directly readable. Neither approach degrades drafting quality.

This unconstrained subspace explains why the 780M MTP states vary sharply across seeds. In every 780M model, a single coordinate of the top-layer hidden state carries over half of the total energy after RMSNorm, yet the final norm assigns it a near-zero gain, rendering it invisible to the output head. Consequently, the MTP loss leaves this coordinate unconstrained in the head states: Seed 1 happens to match the model's true value, Seed 2 flips its sign in Head 1, and Seed 0 leaves it near zero in Heads 1 and 2 while concentrating energy on a different coordinate. However, the feedback path reads $\rms(a)$ without the final norm's gain, turning any mismatch at this coordinate into a large error for the reader. To verify this causally, we clamp this coordinate and the stand-in's largest coordinate to their empirical means from the model's states. The cross-seed error spread drops from roughly 40-fold to 2-fold, though all seeds still incur higher error than a zero-vector baseline.

\begin{table}[H]
\centering\footnotesize\setlength{\tabcolsep}{2.4pt}
\caption{Each seed's drafting quality beside its one-call error $\epsilon^1_a$ and exact decoding BPB on Paloma. A head's acceptance is one minus the total variation between its draft and the exact next-token distribution, teacher-forced, and head $k$ is scored only where heads 1 to $k-1$ were accepted.}
\label{tab:acceptance}
\begin{tabular}{lcccccccccccc}
\toprule
 & & \multicolumn{5}{c}{135M} & \multicolumn{5}{c}{780M} \\
\cmidrule(lr){3-7}\cmidrule(lr){8-12}
Setting & Seed & Head 1 $\uparrow$ & Head 2 $\uparrow$ & Head 3 $\uparrow$ & $\epsilon^1_a$ $\downarrow$ & Exact BPB $\downarrow$ & Head 1 $\uparrow$ & Head 2 $\uparrow$ & Head 3 $\uparrow$ & $\epsilon^1_a$ $\downarrow$ & Exact BPB $\downarrow$ \\
\midrule
MTP state & 0 & 0.730 & 0.721 & 0.715 & 0.0094 & 0.9946 & 0.757 & 0.742 & 0.733 & 1.51\phantom{00} & 0.8288 \\
 & 1 & 0.732 & 0.720 & 0.715 & 0.0047 & 0.9956 & 0.757 & 0.741 & 0.732 & 0.0146 & 0.8291 \\
 & 2 & 0.731 & 0.721 & 0.717 & 0.0065 & 0.9948 & 0.757 & 0.742 & 0.734 & 0.37\phantom{00} & 0.8286 \\
\cmidrule{1-12}
state alignment & 0 & 0.731 & 0.721 & 0.714 & 0.0016 & 0.9948 & 0.757 & 0.741 & 0.732 & 0.0017 & 0.8286 \\
 & 1 & 0.731 & 0.721 & 0.716 & 0.0017 & 0.9950 & 0.758 & 0.741 & 0.732 & 0.0015 & 0.8287 \\
 & 2 & 0.731 & 0.721 & 0.717 & 0.0017 & 0.9953 & 0.757 & 0.742 & 0.733 & 0.0018 & 0.8286 \\
\cmidrule{1-12}
in-place & 0 & 0.730 & 0.720 & 0.716 & 0.0006 & 0.9980 & 0.756 & 0.741 & 0.732 & 0.0006 & 0.8305 \\
 & 1 & 0.730 & 0.720 & 0.717 & 0.0007 & 0.9970 & 0.755 & 0.741 & 0.733 & 0.0006 & 0.8307 \\
 & 2 & 0.729 & 0.719 & 0.716 & 0.0006 & 0.9974 & 0.756 & 0.742 & 0.732 & 0.0006 & 0.8307 \\
\cmidrule{1-12}
in-place, half & 0 & --- & --- & --- & --- & --- & 0.756 & 0.742 & 0.734 & 0.0010 & 0.8298 \\
 & 1 & --- & --- & --- & --- & --- & 0.757 & 0.741 & 0.732 & 0.0010 & 0.8300 \\
 & 2 & --- & --- & --- & --- & --- & 0.756 & 0.741 & 0.734 & 0.0007 & 0.8297 \\
\bottomrule
\end{tabular}
\end{table}

\section{Uncertainty of Downstream Differences}
\label{app:downstream}
\begin{figure}[!htbp]
\centering
\includegraphics[width=\linewidth]{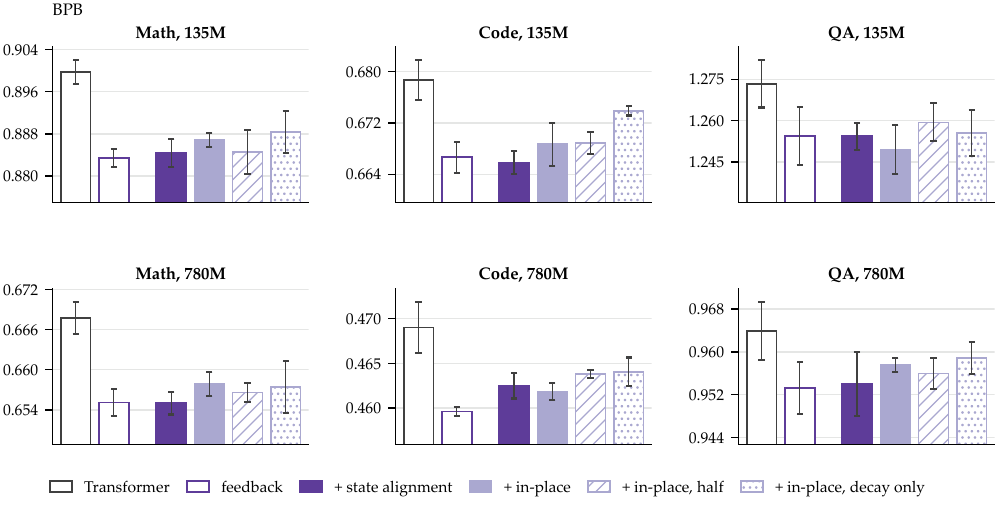}
\caption{Exact decoding BPB of the gold answers in the Olmo~3 Base Easy suites. A + bar's height above the feedback model is that stand-in's ceiling drop $\Delta_a$ (Eq.~\ref{eq:gap}).}
\label{fig:downstream}
\end{figure}

Downstream BPB varies across both training seeds and benchmark items; HumanEval, for instance, contains only 164 problems. Because item-level noise shifts all models similarly and largely cancels out in paired comparisons on identical items, the error bars in Figure~\ref{fig:downstream} reflect seed variance only. Table~\ref{tab:downdiff} reports pairwise differences incorporating both uncertainty sources. The seed variance follows Welch's $t$-test across each model's three seed scores. The item variance is estimated via paired bootstrap as the standard error of the mean difference over 2{,}000 item-level resamples, using identical resampled item sets across both models and all seeds. The two components are summed in variance, with effective degrees of freedom computed via the Welch--Satterthwaite approximation.

\begin{table}[H]
\centering\footnotesize\setlength{\tabcolsep}{4pt}
\caption{Downstream BPB differences, first model minus second, with 95\% intervals over seeds and items. A negative difference favours the first model.}
\label{tab:downdiff}
\begin{tabular}{lccc}
\toprule
Difference & Math & Code & QA \\
\midrule
\multicolumn{4}{l}{\textit{135M}} \\
feedback $-$ Transformer & $-0.0163$ {\scriptsize$[-0.0210, -0.0116]$} & $-0.0121$ {\scriptsize$[-0.0183, -0.0058]$} & $-0.0188$ {\scriptsize$[-0.0409, +0.0033]$} \\
state alignment $-$ feedback & $+0.0009$ {\scriptsize$[-0.0043, +0.0062]$} & $-0.0008$ {\scriptsize$[-0.0056, +0.0040]$} & $-0.0002$ {\scriptsize$[-0.0222, +0.0218]$} \\
in-place $-$ feedback & $+0.0034$ {\scriptsize$[-0.0001, +0.0070]$} & $+0.0020$ {\scriptsize$[-0.0047, +0.0087]$} & $-0.0050$ {\scriptsize$[-0.0273, +0.0173]$} \\
state alignment $-$ in-place & $-0.0025$ {\scriptsize$[-0.0078, +0.0028]$} & $-0.0028$ {\scriptsize$[-0.0094, +0.0038]$} & $+0.0048$ {\scriptsize$[-0.0132, +0.0228]$} \\
\cmidrule{1-4}
\multicolumn{4}{l}{\textit{780M}} \\
feedback $-$ Transformer & $-0.0127$ {\scriptsize$[-0.0177, -0.0077]$} & $-0.0094$ {\scriptsize$[-0.0150, -0.0037]$} & $-0.0107$ {\scriptsize$[-0.0223, +0.0009]$} \\
state alignment $-$ feedback & $-0.0001$ {\scriptsize$[-0.0043, +0.0041]$} & $+0.0028$ {\scriptsize$[+0.0002, +0.0055]$} & $+0.0008$ {\scriptsize$[-0.0117, +0.0133]$} \\
in-place $-$ feedback & $+0.0027$ {\scriptsize$[-0.0016, +0.0070]$} & $+0.0022$ {\scriptsize$[+0.0001, +0.0043]$} & $+0.0043$ {\scriptsize$[-0.0063, +0.0149]$} \\
state alignment $-$ in-place & $-0.0028$ {\scriptsize$[-0.0068, +0.0012]$} & $+0.0006$ {\scriptsize$[-0.0023, +0.0035]$} & $-0.0035$ {\scriptsize$[-0.0170, +0.0101]$} \\
\bottomrule
\end{tabular}
\end{table}

\section{Quality Served by Speculative Decoding}
\label{app:served}
\begin{figure}[!htbp]
\centering
\includegraphics[width=\linewidth]{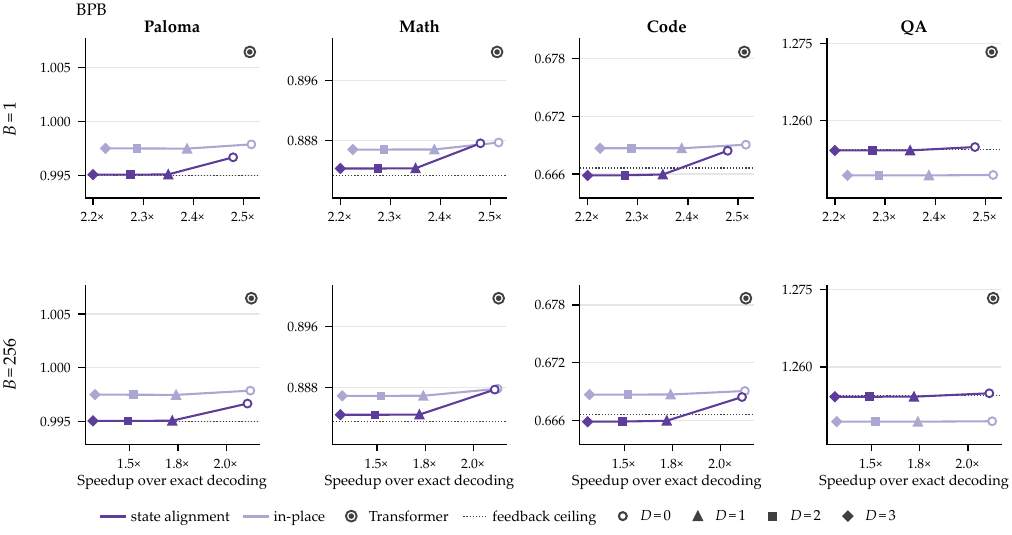}
\caption{Figure~\ref{fig:frontier} at 135M.}
\label{fig:frontier135}
\end{figure}

Figure~\ref{fig:downstream} scores each model's exact decoding, its ceiling. To score what speculative decoding serves, each gold answer is decoded along its own text (Appendix~\ref{app:eval}). Table~\ref{tab:downserved} splits the downstream gap to the feedback ceiling as Eq.~\ref{eq:gap} does. At $D=0$, state alignment adds three to seven times in-place substitution's approximation error on math and code, and both add less than 0.001 on QA; from $D=1$ on, both serve their own ceiling within $10^{-4}$, leaving only the ceiling drop. Figure~\ref{fig:frontier135} repeats Figure~\ref{fig:frontier} at 135M.

\begin{table}[!htbp]
\centering\footnotesize\setlength{\tabcolsep}{3pt}
\caption{The gap to the feedback ceiling on the gold answers of the downstream suites, split as in Eq.~\ref{eq:gap}, in BPB. Drop is the ceiling drop $\Delta_a$ of Figure~\ref{fig:downstream}, whose intervals Table~\ref{tab:downdiff} gives. Error is the approximation error $\epsilon_a$ after $D$ corrections, with 95\% intervals over seeds and items of at most $\pm0.0008$ at $D=0$ and $\pm0.0002$ at $D=1$.}
\label{tab:downserved}
\begin{tabular}{lccccccccc}
\toprule
 & \multicolumn{3}{c}{Math} & \multicolumn{3}{c}{Code} & \multicolumn{3}{c}{QA} \\
\cmidrule(lr){2-4}\cmidrule(lr){5-7}\cmidrule(lr){8-10}
 & & \multicolumn{2}{c}{Error $\downarrow$} & & \multicolumn{2}{c}{Error $\downarrow$} & & \multicolumn{2}{c}{Error $\downarrow$} \\
\cmidrule(lr){3-4}\cmidrule(lr){6-7}\cmidrule(lr){9-10}
Setting & Drop $\downarrow$ & $D{=}0$ & $D{=}1$ & Drop $\downarrow$ & $D{=}0$ & $D{=}1$ & Drop $\downarrow$ & $D{=}0$ & $D{=}1$ \\
\midrule
\multicolumn{10}{l}{\textit{135M}} \\
state alignment & $0.0009$ & $0.0033$ & $0.0000$ & $-0.0008$ & $0.0026$ & $0.0001$ & $-0.0002$ & $0.0007$ & $0.0000$ \\
in-place & $0.0034$ & $0.0009$ & $0.0000$ & $0.0020$ & $0.0004$ & $0.0000$ & $-0.0050$ & $0.0001$ & $0.0000$ \\
\cmidrule{1-10}
\multicolumn{10}{l}{\textit{780M}} \\
state alignment & $-0.0001$ & $0.0036$ & $0.0001$ & $0.0028$ & $0.0024$ & $0.0000$ & $0.0008$ & $0.0004$ & $0.0000$ \\
in-place & $0.0027$ & $0.0013$ & $0.0000$ & $0.0022$ & $0.0004$ & $0.0000$ & $0.0043$ & $0.0001$ & $0.0000$ \\
\bottomrule
\end{tabular}
\end{table}

\end{document}